\documentclass{article}

\usepackage[main, final]{neurips_2026}

\usepackage[utf8]{inputenc} 
\usepackage[T1]{fontenc}    
\usepackage{hyperref}       
\usepackage{url}            
\usepackage{booktabs}       
\usepackage{amsfonts}       
\usepackage{nicefrac}       
\usepackage{microtype}      
\usepackage{xcolor}         
\usepackage{booktabs}
\usepackage{wrapfig}
\usepackage{algorithm}
\usepackage{algorithmic}
\usepackage{color}
\usepackage{colortbl}
\usepackage[svgnames]{xcolor}
\usepackage{amsmath,amssymb,amsfonts}
\usepackage{gensymb}
\usepackage{textcomp}
\usepackage{xspace}
\usepackage{bbding}
\usepackage{multirow}
\usepackage{graphicx}
\usepackage{booktabs}
\usepackage{comment}
\usepackage{array}
\usepackage{pifont}
\newcommand{\cmark}{\ding{51}}   
\newcommand{\xmark}{\ding{55}} 

\title{MemReg: Streaming Outdoor LiDAR Point Cloud Registration with Hybrid Memory Buffers}

\author{%
  Jiuming~Liu\textsuperscript{1}\thanks{Equal Contribution.}, \ \
  Mengmeng~Liu\textsuperscript{2}\footnotemark[1], \ \
  Guangming~Wang\textsuperscript{3}, \ \
  Lihao~Liu\textsuperscript{3}, \ \
  Michael Ying Yang\textsuperscript{4},  \\
  \bf Yunpeng Zhang\textsuperscript{5}, \ \
  Francesco Nex\textsuperscript{2}, \ \
  Hao Cheng\textsuperscript{2}, \ \
  Hesheng Wang\textsuperscript{1}\thanks{Corresponding Author.}\\
  {\textsuperscript{\rm 1} School of Automation and Intelligent Sensing, Shanghai Jiao Tong University}\\
 {\textsuperscript{\rm 2} University of Twente}\quad
 {\textsuperscript{\rm 3} University of Cambridge}\quad
 {\textsuperscript{\rm 4} University of Bath}\quad
 {\textsuperscript{\rm 5} Phigent Robotics}\\
  \texttt{liujiuming123@gmail.com \quad m-liu.1@utwente.nl \quad wanghesheng@sjtu.edu.cn} \\
}

\begin{document}

\maketitle

\begin{abstract}
    Prior outdoor LiDAR registration works are commonly limited by the pair-wise input paradigm, which neglects temporal correlations inherent within streaming sequences. In this paper, we propose a novel multi-frame outdoor point cloud registration network for streaming LiDAR scans enhanced with pose-related memory buffers. The key observation is that long-term temporal LiDAR sequences can provide rich global contextual information to complete sparse measurements, filter outliers, and address low-overlap problems, thereby boosting registration performance. Specifically, two hybrid memory buffers are designed, including an implicit memory feature buffer and an explicit memory pose buffer, to store and dynamically update pose-related temporal features. Moreover, a novel dynamic history weighting module is developed to adaptively fuse current and history pose-related features. Extensive experiments on three outdoor datasets, including KITTI, nuScenes, and Apollo-Southbay, demonstrate state-of-the-art performance of MemReg, surpassing all previous pair-wise methods and also multi-frame SLAM systems. Our method also generalizes surprisingly well to multiview indoor registration scenarios with rather competitive performance on 3DMatch, 3DLoMatch, and ScanNet. Code will be released at \url{https://github.com/IRMVLab/MemReg}.

\end{abstract}    
\section{Introduction}
\label{sec:intro}
Point cloud registration is a fundamental task in the computer vision field, which aims to find the optimal transformation matrix between point cloud frames \cite{li2026combiner,Li_2026_CVPR}. It is widely applied in various downstream tasks, such as autonomous driving \cite{lu2021hregnet,zhang2026diffusion,liu2026driveva,jiang2026vectorworld,liu2026universe,liu2023tracing,dong2026languageconditionedworldmodelingvisual,dong2026towards,dong2026havln20openbenchmark} and SLAM systems \cite{liu2023translo,deng2024plgslam,liu2026streamvlo,zhu2024sni,liu2025dvlo4d,deng2026voxelized}.

Even though object-level and indoor point cloud registration have been widely explored \cite{huang2021predator,qin2022geometric,yew2022regtr,ao2023buffer}, the large-scale outdoor LiDAR registration is still underexplored \cite{lu2021hregnet,liu2023regformer}. Challenges are mainly three-fold: (1) Distinct from object-level point cloud, outdoor points are typically more sparse, irregular, noisy, and have a wider spatial distribution range \cite{lu2023hregnet}. 
(2) Outliers, e.g., occlusions and highly dynamic objects, introduce inconsistent motion patterns, undermining the final ego-motion regression \cite{liu2023regformer,liu2024difflow3d}.
(3) Low-overlap point pairs pose another significant challenge \cite{huang2021predator}. Due to the high speed of ego vehicles in autonomous driving scenarios, consecutive LiDAR scans often have large translational displacements, making it difficult to search for neighboring points across frames \cite{xue2024hdmnet}. To tackle these issues, HRegNet \cite{lu2021hregnet} addresses mismatches and ensures reliable correspondences caused by outliers with bilateral and neighborhood consensus in point-matching layers. RegFormer \cite{liu2023regformer} uses a projection-aware Transformer with enlarged receptive fields to handle the low-overlap issue. \emph{However, these prior outdoor registration methods only take pair-wise point clouds as inputs, ignoring intrinsically rich temporal information in streamlined LiDAR sequences.}

Inspired by recent advances in developing temporal cues for tasks like semantic segmentation \cite{li2023memoryseg,lao2023simultaneously,chen2023svqnet,zheng2024spherical}, object detection \cite{wang2023exploring,hou2024query}, and multi-object tracking \cite{gao2023memotr,chen2026egoadapt}, we aim to improve registration accuracy by extending pair-wise frames to streaming multi-frame inputs. The key motivations are: (1) Long-term temporal information helps complete scenes by supplementing sparse and noisy observations across consecutive frames. (2) Multi-frame inputs enhance temporal consistency, improving the recognition and rejection of outliers for effective point correlation. (3) For the low-overlap issue, motion priors from past timestamps can provide strong pose initialization and reduce the distance between initially distant point pairs. These advantages are demonstrated in Sec. \ref{sec:diss}.

\begin{wrapfigure}{r}{0.68\textwidth}
  \vspace{-10pt}  
  \centering
  \includegraphics[width=1\linewidth]{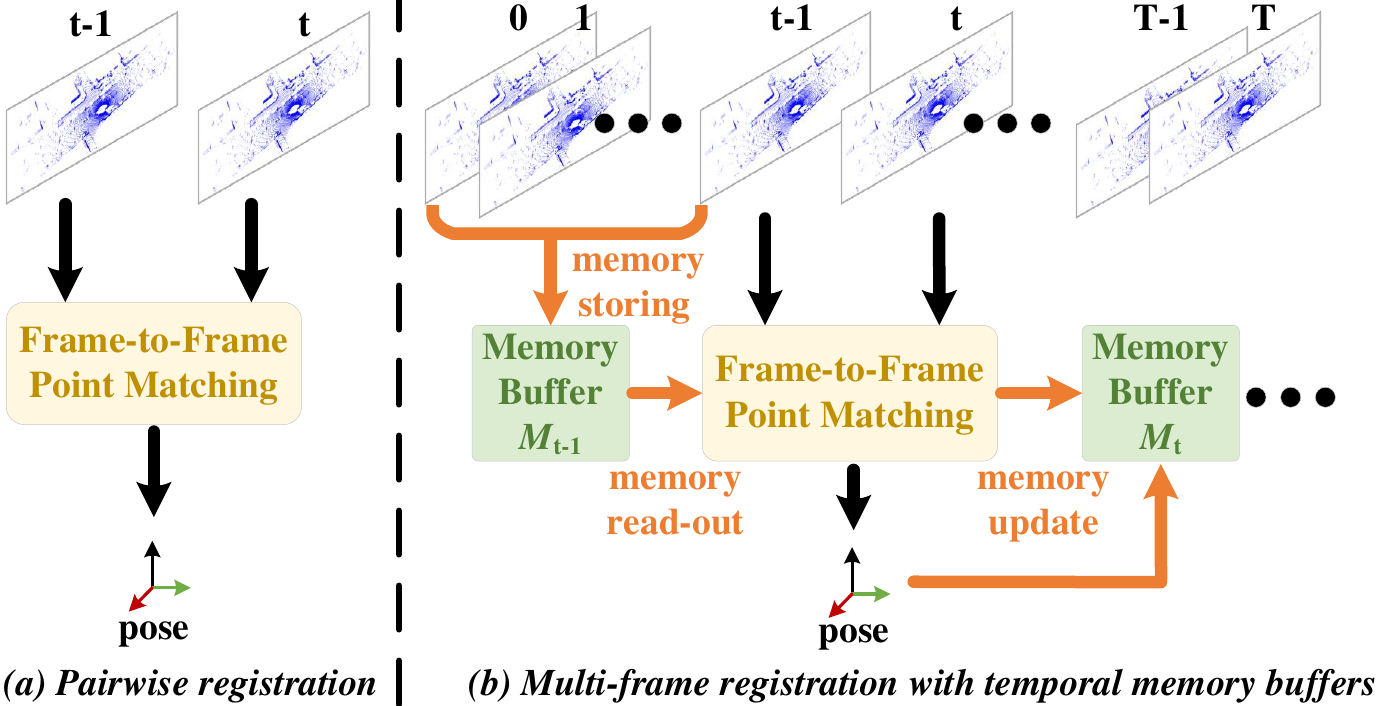}
\vspace{-6mm}
\caption{\textbf{Comparison with prior pair-wise registration methods.} 
Prior methods are commonly constrained by pair-wise registration (a), lacking temporal exploration in LiDAR sequences. In contrast, we propose a streaming multi-frame registration network using long-term temporal priors from hybrid memory buffers (b).
}
\vspace{-3mm}
\label{fig:compare}
\end{wrapfigure}

In this paper, we propose a novel streaming multi-frame LiDAR registration network, termed MemReg, designing temporal memory buffers and dynamic history weighting to explore inherent temporal consistency as in Fig.~\ref{fig:compare}. Temporal cues from multiple history frames are progressively stored and retrieved in an autoregressive frame-by-frame manner to improve each pair-wise registration, rather than directly outputting multi-frame poses. Hybrid memory buffers are designed to store implicit temporal features (\emph{Memory Feature Buffer}) and explicit historical transformations (\emph{Memory Pose Buffer}) with a fixed memory length. Specifically, when the current pair of frames is delivered into the network, the \emph{Memory Feature Buffer} is first updated by concatenating the current motion features and pre-stored multi-frame history motion features. Then, a Mamba-based \cite{gu2023mamba,liu20264dstr} temporal encoding layer is developed to establish long-term temporal dependencies across the updated memory features with high efficiency. Similarly, current pose transformations and temporally stored ones are processed by another Mamba module as a strong pose initialization in the \emph{Memory Pose Buffer}, generating transformation features. To effectively merge history and current pose-related features, a dynamic history weighting module inspired by \cite{jia2016dynamic,aydemir2023adapt} is designed to rescale history and current features and adaptively weight current features with multi-frame historical ones. Finally, the current pose is regressed and refined, which is used to update memory buffers for the next timestamp within sequential LiDAR pairs autoregressively.  

Overall, the key contributions of this paper are:
\begin{itemize}
\item We propose a novel outdoor LiDAR point cloud registration paradigm enhanced by multi-frame streaming motions, which improves the robustness to challenging cases like outliers and low-overlap inputs.

\item Explicit and implicit memory buffers store temporal information from historical frames. Memory Feature Buffer holds implicit pose-related features, while Memory Pose Buffer maintains explicit history poses for enhanced pose initialization. 

\item To effectively integrate history and current features, a dynamic history weighting module is designed to adaptively merge features on different scales.

\item Our MemReg achieves state-of-the-art performance on outdoor registration datasets, including KITTI \cite{geiger2012we}, nuScenes \cite{caesar2020nuscenes}, and Apollo-Southbay \cite{lu2019l3}. Furthermore, our method generalizes well to the indoor multiview and pairwise registration tasks on 3DMatch, 3DLoMatch \cite{zeng20173dmatch}, and ScanNet \cite{dai2017scannet} benchmarks. 

\end{itemize}

\section{Related Work}

\noindent\textbf{Outdoor Point Cloud Registration.} Point cloud registration \cite{121791,zeng20173dmatch,deng2018ppfnet,choy2019fully,aoki2019pointnetlk,dong2020registration,qin2022geometric,mu2024colorpcr,jia2024incremental,li2026retrack,li2026tema,chen2024dynamic,yuan2024inlier,jiang2024se,fu2025dual,lu2026worldcoder,gong2024eadreg} is a crucial task that targets at finding the optimal transformation matrix between point cloud frames. Recently, large-scale outdoor registration has raised remarkable research attention, but the original sparsity, irregularity, and high-dynamic characteristics of outdoor LiDAR points pose significant challenges \cite{liu2025difflow3d,jiang20243dsflabelling,liu2025topolidm}. 
As a pioneering work, HRegNet \cite{lu2021hregnet} proposes bilateral consensus and neighborhood consensus to reduce mismatches. RegFormer \cite{liu2023regformer,liu2026regformer++} designs a projection-aware Transformer, incorporating the cross-attention and all-to-all point-gathering strategy for reliable cross-frame correlation. To address density disparities, GCLNet \cite{liu2023density} introduces a group-wise contrastive learning to extract density-invariant geometric features for registration. SDMNet \cite{lu2023sparse} develops a sparse-to-dense registration network with a sparse matching stage and a local-dense matching stage. Darls \cite{wang2025density} designs a density adaptive registration method for large-scale point clouds. To improve generalization, UGP \cite{zeng2025unlocking} proposes inter-frame cross attention and progressive self attention. However, these previous works typically take pair-wise-only point pairs and neglect the inherent temporal consistency within LiDAR sequences. In this paper, we propose a novel outdoor registration model enhanced by streamlined temporal consistency.

\noindent\textbf{Temporal Modeling with Memory.} Sequential 4D point cloud streams have great potential in boosting various dense prediction tasks \cite{wang2024sequential,liu2026towards,liu2024laformer,cheng2023gatraj}. Different from a static point cloud with limited observations, the additional temporal dimension can supplement incomplete measurements \cite{li2023memoryseg}, display consistent motion priors \cite{gao2023memotr}, and further boost the performance of original single-frame tasks \cite{wang2024sequential,fu2026egoaction}. Recently, increasing research has focused on storing temporal information with a memory mechanism. STMM \cite{xiao2018video} develops a recurrent unit with memory storage for object detection, modeling the long-term temporal appearance and motion dynamics. MeMOT \cite{cai2022memot} designs a memory-based multi-object tracking framework that utilizes a long-term spatio-temporal memory storing identity features of tracked objects. MemorySeg \cite{li2023memoryseg} proposes a 3D latent memory representation to improve current predictions with past frames. MEMFOF \cite{Bargatin_2025_ICCV} and ARFlow \cite{liu2026arflow} present memory-enhanced optical flow pipeline based on high-resolution predictions. Inspired by these works, we design both explicit and implicit memory buffers for temporal information storage.

\section{Method}
\subsection{Overall Architecture}
\begin{figure*}[t]
\centering
\includegraphics[width=1.0\linewidth]{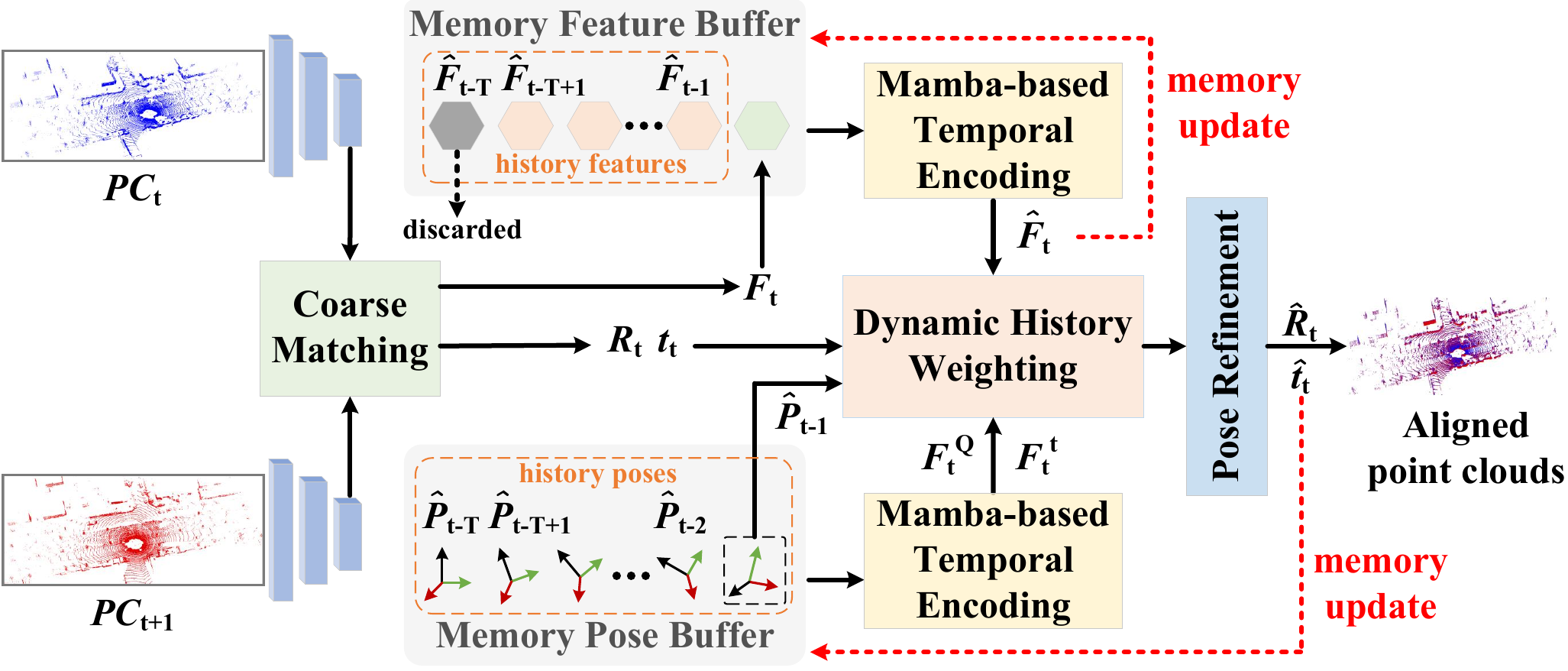}
\vspace{-6mm}
\caption{\textbf{The overview architecture of MemReg.} We leverage the long-term temporal features from history frames to enhance the current registration. During the current registration, memory features and memory poses are first read-out and interacted with current motion features through the Mamba-based Temporal Encoding module and Dynamic History Weighting module. Then, the current refined poses will, in turn, update memory buffers for next-timestamp estimation.}
\vspace{-3mm}
\label{fig:pipeline}
\end{figure*}

Given a stream of LiDAR point cloud sequences $\{PC_{0}, PC_{1}, ..., PC_{\mathrm{V-1}}\}^{\mathrm{V}\times\mathrm{N}\times3}$, we propose a multi-frame registration framework as described in Fig. \ref{fig:pipeline}, where buffers storing multiple history memories would enhance the performance of each pair-wise registration. Specifically, to correlate current point pairs $PC_\mathrm{t}$ and $PC_\mathrm{t+1}$, hierarchical point features are first extracted, and then matched in the coarsest layer following HRegNet \cite{lu2023hregnet}. Afterwards, history motion information is read-out from memory buffers on both feature level and pose level in Sec. \ref{sec:mem_out}. To dynamically merge history and current features, we also design a dynamic history weighting module in Sec. \ref{sec:weighting}. Finally, estimated coarse poses are iteratively refined, generating the final rotation matrix $\hat{R}\in\mathbb{R}^{3\times3}$ and translation vector $\hat{t}\in\mathbb{R}^{3\times1}$. Memory buffers are also dynamically updated with current features and current poses in Sec. \ref{sec:mem_update}. The network is supervised by the loss function defined in Sec. \ref{sec:loss}.

\subsection{Memory Initialization and Read-out}
\label{sec:mem_out}
Unlike pair-wise registration, streaming point cloud videos are the inputs of our network for long-term temporal modeling. To take advantage of temporal motion priors sufficiently, we design two memory buffers storing both cross-frame implicit feature-level memory and explicit pose-level memory. Our memory buffers $M^\mathrm{F}_{0}\in\mathbb{R}^\mathrm{T\times D_f}$ of features and $M^\mathrm{P}_{0}\in\mathbb{R}^\mathrm{T\times D_p}$ of poses are initialized by empty sets, where the temporal length of both memory buffers is maintained as $T$.

\vspace{3pt}
\noindent\textbf{Implicit Feature Memory Read-out.} With motion features $F_\mathrm{t}$ and poses $R_\mathrm{t}$, $t_\mathrm{t}$ extracted from the coarse matching layer in the current timestamp, we first read-out the temporally-stored features using a sliding window mechanism. When current motion features are delivered into the Memory Feature Buffer $M^\mathrm{F}_\mathrm{t-1}$, the temporally-farthest feature in the buffer $\hat{F}_\mathrm{t-T-1}$ is discarded. Other $T-1$ temporally-nearest features $\{\hat{F}_\mathrm{t-T},..., \hat{F}_\mathrm{t-1}\}$ are concatenated with the current motion feature $F_\mathrm{t}$ as:
\begin{equation}
    \label{eq:concat}
     M^\mathrm{F}_\mathrm{t-1}: \{\hat{F}_\text{t-T-1},\hat{F}_\text{t-T+1},...,\hat{F}_\text{t-1}\} \Rightarrow \{\hat{F}_\text{t-T}, ..., \hat{F}_\text{t-1},F_\text{t}\},
     \vspace{-2pt}
\end{equation}
where \{\} indicates feature concatenation along the temporal dimension. The concatenated features formulate the input tokens of the following Mamba-based Temporal Encoding \cite{gu2023mamba,liu2025mamba4d}:
\begin{equation}
    \label{eq:mamba1}
     \hat{M}^\mathrm{F}_\mathrm{t-1} = \text{LN}( M^\mathrm{F}_\mathrm{t-1}),
\end{equation}
\begin{equation}
    \label{eq:mamba2}
     \overline{M}^\mathrm{F}_\mathrm{t} = \delta(\text{DW}(\text{Linear}(\hat{M}^\mathrm{F}_\mathrm{t-1}))),
\end{equation}
\begin{equation}
    \label{eq:mamba3}
     \hat{M}^\mathrm{F}_\mathrm{t} = \delta(\text{Linear}(\hat{M}^\mathrm{F}_\mathrm{t-1})),
\end{equation}
\begin{equation}
    \label{eq:mamba4}
     \hat{F}_\text{t} = \text{Linear}(\text{SSM}(\overline{M}^\mathrm{F}_\mathrm{t}))\odot\hat{M}^\mathrm{F}_\mathrm{t} + M^\mathrm{F}_\mathrm{t-1},
     \vspace{-2pt}
\end{equation}
where $\delta$ denotes the SiLU activation \cite{hendrycks2016gaussian}. DW is the depth-wise convolution \cite{chollet2017xception}. LN indicates the Layer Normalization. SSM indicates the standard selective state space model \cite{gu2023mamba}. The output $\hat{F}_{t}$ is the temporally encoded feature. 

\vspace{3pt}
\noindent\textbf{Explicit Pose Memory Read-out.} $T$-length nearest Memory Pose Buffer $M^\mathrm{P}_\mathrm{t-1}$ consists of historically estimated rotation matrices $M^\mathrm{R}_\mathrm{t-1} = \{\hat{Q}_\text{t-T-1},\hat{Q}_\text{t-T},...,\hat{Q}_\text{t-1}\}$ and translation matrices $M^\mathrm{t}_\mathrm{t-1} = \{\hat{t}_\text{t-T-1},\hat{t}_\text{t-T},...,\hat{t}_\text{t-1}\}$, which are respectively embedded by two separate MLP layers first to lift their dimensions. To facilitate regression, the rotation matrix here is represented by quaternion vectors $\hat{Q}_\text{t}\in\mathbb{R}^{4\times1}$. Finally, the embedded rotation features and translation features are fed into two Mamba blocks for the temporal modeling, respectively:
\begin{equation}
    \label{eq:mamba5}
     F^{Q}_{t}= \text{Mamba}(\text{MLP}(\{\hat{Q}_\text{t-T-1},\hat{Q}_\text{t-T},...,\hat{Q}_\text{t-1}\})),
\end{equation}
\begin{equation}
    \label{eq:mamba6}
     F^{t}_{t}= \text{Mamba}(\text{MLP}(\{\hat{t}_\text{t-T-1},\hat{t}_\text{t-T},...,\hat{t}_\text{t-1}\})),   
\end{equation}
where $\hat{Q}_\text{t}$ and $\hat{t}_\text{t}$ respectively denote the $t$-th temporally-nearest quaternion and translation vectors stored in the Memory Pose Buffer. Here, Mamba indicates the same operation as in Memory Feature Buffer, about which we avoid detailed descriptions for simplicity.


\subsection{Dynamic History Weighting}
\label{sec:weighting}
After retrieving the history-interacted features, the key problem is how to merge history-interacted features with the current motion features for enhanced registration performance. Since there may be scale gaps, the history and current poses should be re-scaled for effective information merging. Inspired by the great success of the dynamic weighting mechanism in \cite{jia2016dynamic,aydemir2023adapt}, a dynamic history weighting layer is introduced to adaptively merge the history pose-related features, history poses, current pose-related features, and current poses. 

To be specific, the inputs of our dynamic history weighting layer are composed of the history pose $\hat{Q}_{t-1}, \hat{t}_{t-1}$ from the latest previous timestamp, history pose-encoded features $F^{Q}_{t}, F^{t}_{t}$, current pose $R_{t}, t_{t}$, and current pose-encoded features $\hat{F}_{t}$. The current rotation matrix $R_{t}$ is also converted to the quaternion vector $Q_{t}$ for consistency. History and current pose-related features are first concatenated:
\begin{equation}
    \label{eq:feature gathering_2}
     F^{Q}_{d} = \hat{Q}_{t-1} \oplus F^{Q}_{t} \oplus Q_{t} \oplus \hat{F}_{t},
\end{equation}
where $\oplus$ indicates concatenation along the channel dimension. Then, the dynamic weighting head is designed as:
\begin{equation}
      \label{eq:head}
          W_{1}=W_{d_{1}} F^{Q}_{d},\quad W_{2} = W_{d_{2}} F^{Q}_{d}, 
\end{equation}
\begin{equation}
      \label{eq:head2}
         F^{Q}_{d_{1}}= \mathrm{ReLU}(\mathrm{Norm}(W_{1} F^{Q}_{d})), \quad
          Q^{d}_{t} = W_{2} F^{Q}_{d_{1}}, 
\end{equation}
where $W_{d_{1}}$ and $W_{d_{2}}$ are trainable parameters, and the output dimension is set to the same as the quaternion vector. Here, we only give descriptions of the quaternion vector weighting, and the same applies to the translation matrix weighting, generating the weighted translation vector $t^{d}_{t}$. By providing history pose-related features to the dynamic prediction head as inputs, the learnable weights enable the network to adjust to different scales dynamically and re-weight current pose estimations with history motion priors.

\subsection{Memory Update and Pose Refinement}
\label{sec:mem_update}
Our fixed-length memory buffers follow the sliding window mechanism, which are progressively updated as streaming point frames are delivered sequentially into the network. For the Memory Feature Buffer update, we directly concatenate the encoded current features $\hat{F}_{t}$ with $T$-1 previous features:
\begin{equation}
    \label{eq:concat2}
     M^\mathrm{F}_\mathrm{t} = \{\hat{F}_\text{t-T+1}, ..., \hat{F}_\text{t-1}, \hat{F}_\text{t}\}.
\end{equation}
For the Memory Pose Buffer update, the pose refinement in \cite{lu2023hregnet} is first adopted to refine the poses $Q^{d}_{t}$ and $t^{d}_{t}$ to get the final pose estimation $\hat{R}_{t}$ and $\hat{t}_{t}$. The Memory Pose Buffer is then updated by concatenating previous $T$-1 poses and the current pose estimate as:
\begin{equation}
    \label{eq:concat3}
     M^\mathrm{R}_\mathrm{t} = \{\hat{R}_\text{t-T+1}, ..., \hat{R}_\text{t-1}, \hat{R}_{t}\}, \quad
     M^\mathrm{t}_\mathrm{t} = \{\hat{t}_\text{t-T+1}, ..., \hat{t}_\text{t-1}, \hat{t}_{t}\}.
\end{equation}

\subsection{Loss Function}
\label{sec:loss}
Our network is supervised by reducing the discrepancy for each pair-wise registration sample \cite{lu2023hregnet,liu2023regformer}. The rotation and translation loss functions are:
\begin{equation}
    \label{eq:loss}
     \mathcal{L}_\text{rot} = || \hat{R}^{T}R - \text{I} ||_{2},  \quad
          \mathcal{L}_\text{trans} = || \hat{t} -t ||_{2},
\end{equation}
where $\hat{R}, \hat{t}$ indicate the estimated rotation and translation matrices. $R, t$ are the ground truth rotation and translation matrices. I denotes the identity matrix.

The overall loss function is: $\mathcal{L} = \mathcal{L}_\text{trans} +  \mathcal{L}_\text{rot}.$

\section{Experiment}
\subsection{Experimental Settings}
\textbf{Datasets and Evaluation Metrics.} We conduct outdoor evaluations on three commonly used LiDAR datasets: KITTI odometry \cite{geiger2012we}, nuScenes \cite{caesar2020nuscenes}, and Apollo-Southbay \cite{lu2019l3}. To further demonstrate the generalization ability of our method, we also utilize indoor datasets including ScanNet \cite{dai2017scannet}, 3DMatch, and 3DLoMatch \cite{zeng20173dmatch} for evaluation. Following the protocols in previous works \cite{lu2023hregnet,lu2023sparse,liu2023regformer}, we use three main metrics to assess our method: Relative Translation Error (RTE), Relative Rotation Error (RRE), and Registration Recall. 
A successful registration is decided only when the RTE and RRE are both within certain thresholds $\theta_{T}, \theta_{R}$. To avoid the unreliable error metric report by minor failure samples, we follow previous works \cite{lu2021hregnet,xue2024hdmnet,liu2023regformer} to calculate the average RTE and RRE only for successful registration samples.

\setlength{\tabcolsep}{0.3mm}
\begin{table}[t]
\begin{minipage}{0.485\textwidth}
   \centering
 \caption{\textbf{Outdoor point cloud registration on the KITTI dataset \cite{geiger2012we}.} `*' indicates the baseline on which we introduce multi-frame memory.}
	\label{tab:reg}
    \vspace{-6pt}
	\footnotesize
    \resizebox{1.00\columnwidth}{!}
            {
	\begin{tabular}{l|cccc}
        \hline
		\toprule
		\multirow{2}{*}{Methods}&\multicolumn{4}{|c}{KITTI Odometry}\\
		\cmidrule(r){2-5}
		~ & RTE (m) & RRE (deg) & Recall &Time (ms) \\
		\midrule

		ICP (P2P) \cite{121791} & $0.045\pm 0.054$ & $0.112\pm 0.093$ & 14.25\% &472.2\\
		ICP (P2Pl) \cite{121791} & $0.044\pm 0.041$ & $0.145\pm 0.153$ & 33.56\%&461.7 \\
        RANSAC \cite{fischler1981random} & $0.126\pm 0.067$ & $0.538\pm 0.396$ & 91.94\% &459.4 \\
        FGR \cite{zhou2016fast} & $0.929\pm 0.592$ & $0.963\pm 0.806$ & 39.42\% &506.1 \\
	
		\midrule
		IDAM \cite{li2020iterative} & $0.659\pm 0.483$ & $1.057\pm 0.939$ & 70.92\% & 33.4\\
		FMR \cite{huang2020feature} & $0.657\pm 0.422$ & $1.493\pm 0.847$ & 90.57\%& 85.5\\ 
		DGR \cite{choy2020deep} & $0.322\pm 0.319$ & $0.374\pm 0.302$ & 98.71\% &1496.6  \\
        HRegNet \cite{lu2021hregnet} & $0.056\pm 0.075$ & $0.178\pm 0.196$ & 99.77\% &106.2\\
        RegFormer \cite{liu2023regformer} & $0.082\pm 0.112$ & $0.228\pm 0.209$ & 99.80\% & 98.3\\
        OKR-Net \cite{wang2023okr} & $0.070\pm 0.130$ & $0.230\pm 0.220$ & 99.22\%  &100.4\\
        HRegNet* \cite{lu2023hregnet} & $0.047\pm 0.037$ & $0.147\pm 0.120$ & \textbf{100.0}\% &136.0\\
        HDMNet \cite{xue2024hdmnet} & $0.050\pm 0.057$ & $0.159\pm 0.152$ & 99.85\% & 120.2\\
         PTT \cite{wang2025point}& \hspace{-32pt}$0.063$ & $0.230$\hspace{32pt} & 99.80\% & -\\

        UPG \cite{zeng2025unlocking} & \hspace{-32pt}$0.071$ & $0.245$\hspace{32pt} & 99.80\% &-\\
         
        LSReg-Net \cite{tao2025lsreg} & $0.060\pm 0.070$ & $0.200\pm 0.170$ & 99.92\% &92.5\\

        \midrule
        Ours & $\bf{0.032\pm 0.036}$ & $\bf{0.122\pm 0.096}$ & \textbf{100.0}\% & 133.0\\
		\bottomrule\hline
	\end{tabular}
 }
\end{minipage}
\hspace{3pt}
\begin{minipage}{0.485\textwidth}
   \centering
\caption{\textbf{Outdoor point cloud registration on the nuScenes dataset \cite{caesar2020nuscenes}.} `*' indicates the baseline on which we introduce multi-frame memory.}
	\label{tab:reg2}
    \vspace{-6pt}
	\footnotesize
    \resizebox{1.00\columnwidth}{!}
            {
	\begin{tabular}{l|cccc}
        \hline
		\toprule
		\multirow{2}{*}{Methods}&\multicolumn{4}{|c}{nuScenes}\\
		\cmidrule(r){2-5}
		~ & RTE (m) & RRE (deg) & Recall &Time (ms) \\
		\midrule
        
		ICP (P2P) \cite{121791}  & $0.252\pm 0.510$ & $0.253\pm 0.502$ & 18.78\%&82.0  \\
		ICP (P2Pl) \cite{121791}  & $0.153\pm 0.296$ & $0.212\pm 0.306$ & 36.8\% &44.5 \\
        RANSAC \cite{fischler1981random}   & $0.206\pm 0.186$ & $0.738\pm 0.704$ & 60.91\%&187.4  \\
        FGR \cite{zhou2016fast}   & $0.708\pm 0.622$ & $1.007\pm 0.924$ & 32.2\% &284.6 \\
		
		\midrule
		DCP \cite{wang2019deep}  & $1.087\pm 0.491$ & $2.065\pm 1.141$ & 58.58\% &45.5\\
        IDAM \cite{li2020iterative} & $0.467\pm 0.410$ & $0.793\pm 0.783$ & 87.98\% &32.6 \\
        FMR \cite{huang2020feature}  & $0.603\pm 0.391$ & $1.610\pm 0.974$ & 92.06\%&61.1  \\ 
	      DGR \cite{choy2020deep}  & $0.211\pm 0.183$ & $0.476\pm 0.430$ & 98.41\% &523.0\\
        HRegNet \cite{lu2021hregnet}   & $0.122\pm 0.112$ & $0.273\pm \bf0.197$ & \textbf{100.0}\% &87.3 \\
        RegFormer \cite{liu2023regformer}   & $0.198$ \hspace{28pt} & $0.223$ \hspace{28pt} & 99.90\% &85.6 \\
        OKR-Net \cite{wang2023okr}  & $0.140\pm 0.130$ & $0.280\pm 0.210$ & 99.96\% &60.0 \\
        HRegNet* \cite{lu2023hregnet}   & $0.110\pm 0.096$ & $0.285\pm 0.209$ & \textbf{100.0}\%&120.1  \\
		HDMNet \cite{xue2024hdmnet}  & $0.114\pm 0.102$ & $0.274\pm 0.206$ & \textbf{100.0}\%&102.9  \\
		
        UPG \cite{zeng2025unlocking} & \hspace{-32pt}$0.179$ & $0.270$\hspace{32pt} & \bf 100.0\% &-\\
        
         LSReg-Net \cite{tao2025lsreg} & $0.150\pm 0.130$ & $0.320\pm 0.220$ & 99.98\% &81.2\\

        \midrule
        Ours  & $\bf{0.069\pm 0.055}$ & $\bf{0.261\pm 0.197}$ & \textbf{100.0}\%& 119.3\\
		\bottomrule\hline
	\end{tabular}
 }
\end{minipage}%
\vspace{-2mm}
\end{table}

\begin{figure}[t]
\centering
\includegraphics[width=0.95\linewidth]{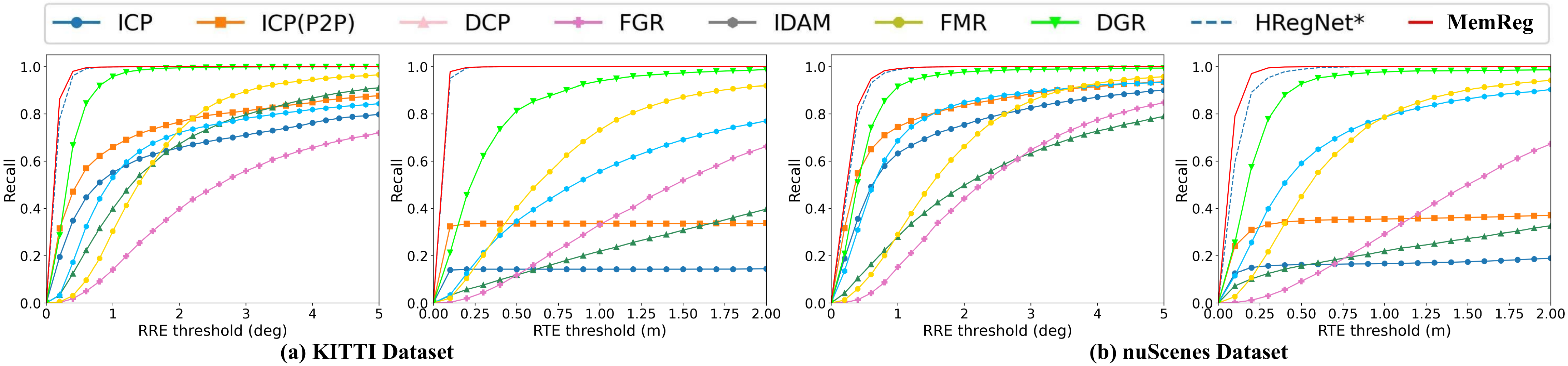}
\vspace{-3mm}
\caption{\textbf{The registration recall under different RRE or RTE thresholds on the KITTI and nuScenes datasets.}}
\vspace{-3mm}
\label{fig:thre}
\end{figure}

\vspace{3pt}
\noindent\textbf{Implementation Details.} We first voxelize the input points with a voxel size of 0.3\,m. Following \cite{lu2021hregnet,lu2023sparse,xue2024hdmnet}, down-sampling is adopted for the input points, where 16,384 points are randomly sampled for KITTI and Apollo-Southbay, and 8,192 points for nuScenes. Experiments are conducted on a single NVIDIA RTX3090 GPU. The Adam optimizer is adopted with $\beta_{1}$ = 0.9, $\beta_{2}$ = 0.999. The initial learning rate is 0.001 and exponentially decays every 200,000 steps until reaching 0.00001. The batch size is set to 8. We choose HRegNet$^{*}$ \cite{lu2023hregnet} here as the baseline, on which we introduce multi-frame temporal information. 
Registration is successful when $\theta_{T}$ = 5m, $\theta_{R}$ = 2\,degree (deg) by default.
The temporal length in memory buffers $T$ is set to 20 for KITTI and Apollo-Southbay, and 10 for nuScenes.

\subsection{Quantitative Results}

\vspace{3pt}
\noindent\textbf{Comparison Results on KITTI.} As in Table \ref{tab:reg}, our model not only outperforms previous optimization methods, e.g., ICP \cite{121791}, RANSAC \cite{fischler1981random}, by a large margin, but also learning-based methods. Compared to the previous SOTA method HRegNet$^{*}$ \cite{lu2023hregnet}, our model achieves a 31.9\% lower average RTE and a 17.0\% lower average RRE enhanced by adding streaming temporal information. Also, compared to the baseline \cite{lu2023hregnet}, our method requires a slightly shorter runtime for one inference. The experiment results demonstrate both the accuracy and efficiency of our proposed method.

\setlength{\tabcolsep}{0.3mm}
\begin{table}[t]
\begin{minipage}{0.485\textwidth}
   \centering
 \caption{\textbf{Outdoor point cloud registration performance on the Apollo-Southbay dataset \cite{lu2019l3}.} `*' indicates the baseline method on which we introduce multi-frame temporal information.}
	\label{tab:reg3}
    \vspace{-6pt}
	\footnotesize
    \resizebox{1.00\columnwidth}{!}
            {
	\begin{tabular}{l|cccc}
        \hline
		\toprule
		\multirow{2}{*}{Methods}&\multicolumn{4}{|c}{Apollo-Southbay}\\
		\cmidrule(r){2-5}
		~ & RTE (m) & RRE (deg) & Recall &Time (ms) \\
		\midrule
        
		ICP (P2P) \cite{121791}  & $0.100\pm 0.039$ & $0.063\pm 0.305$ & 39.93\%&482.1  \\
		ICP (P2Pl) \cite{121791}  & $0.039\pm 0.179$ & $0.046\pm 0.257$ & 46.45\% &470.2 \\
       
        RANSAC \cite{fischler1981random}   & $0.125\pm 0.114$ & $0.361\pm 0.368$ & 83.72\%&552.1  \\
         FGR \cite{zhou2016fast}   & $0.470\pm 0.498$ & $0.664\pm 0.762$ & 50.50\% &496.0 \\
		\midrule
		DCP \cite{wang2019deep}  & $1.174\pm 0.499$ & $2.155\pm 1.254$ & 28.49\% &41.0\\
        IDAM \cite{li2020iterative} & $0.456\pm 0.416$ & $0.361\pm 0.429$ & 74.03\% &32.0 \\
        FMR \cite{huang2020feature}  & $0.653\pm 0.488$ & $0.727\pm 0.658$ & 83.78\%&83.4  \\ 
	DGR \cite{choy2020deep}  & $0.132\pm 0.151$ & $0.127\pm 0.146$ & 99.64\% &2507.0\\
		HRegNet* \cite{lu2023hregnet}   & $0.034\pm 0.037$ & $0.079\pm 0.079$ & 99.88\%&129.5  \\

        \midrule
        Ours  & $\bf{0.024\pm 0.031}$ & $\bf{0.072\pm 0.074}$ & \textbf{100.0}\%& 127.6\\
		\bottomrule\hline
	\end{tabular}
 }
\end{minipage}
\hspace{3pt}
\setlength{\tabcolsep}{0.75mm}
\begin{minipage}{0.485\textwidth}
   \centering
\caption{\textbf{Comparison with SLAM systems on KITTI odometry \cite{geiger2012we}.} We report the average sequence translational RMSE (\%). `Multi.' indicates methods modeling multi-frame correlation.}
	\label{tab:reg4}
    \vspace{-6pt}
	\footnotesize
   \resizebox{1\columnwidth}{!}{
    \begin{tabular}{l|c|ccccc}
    \hline\toprule
     Method & Multi. &07 & {08} & {09} & {10} & Mean(07-10) \\
    \midrule
   LONet \cite{li2019net}&\xmark &1.70&2.12&1.37&1.80&1.75\\
   PWCLO \cite{wang2021pwclo} & \xmark&1.16    & 1.68    & 0.88    & 2.14    & 1.47    \\
   TransLO \cite{liu2023translo}& \xmark &0.55&1.29&0.95&1.18&0.99\\
   DVLO \cite{liu2024dvlo} &\xmark& 0.46  & 1.09  & 0.85  & 0.88  & 0.82  \\
   LAGLO \cite{tang2025enhanced} &\xmark &1.02&1.48&2.29&2.00&2.15\\
   \midrule
   LOAM \cite{zhang2014loam}&\cmark&0.69&1.18&1.20&1.51&1.15\\
   DELO~\cite{ali2023delo} &\cmark &0.58  &1.36  &1.23  &1.53  &1.18 \\
   DSLO \cite{zhang2024dslo} &\cmark & 0.58  & 1.16  & \bf0.72  & 1.29  & 0.94 \\
   DVLO4D \cite{liu2025dvlo4d}&\cmark&0.43  & 0.95  & 0.77  & 0.76 & 0.73 \\ 
   Ours & \cmark&\bf0.38  & \bf0.81  & 0.75  & \bf0.72  & \bf0.67  \\
    \bottomrule\hline
    \end{tabular}%
    }
\end{minipage}%
\vspace{-4mm}
\end{table}

\begin{figure*}[t]
\centering
\includegraphics[trim=3pt 0pt 3pt 0pt, clip=true, width=0.95\linewidth]{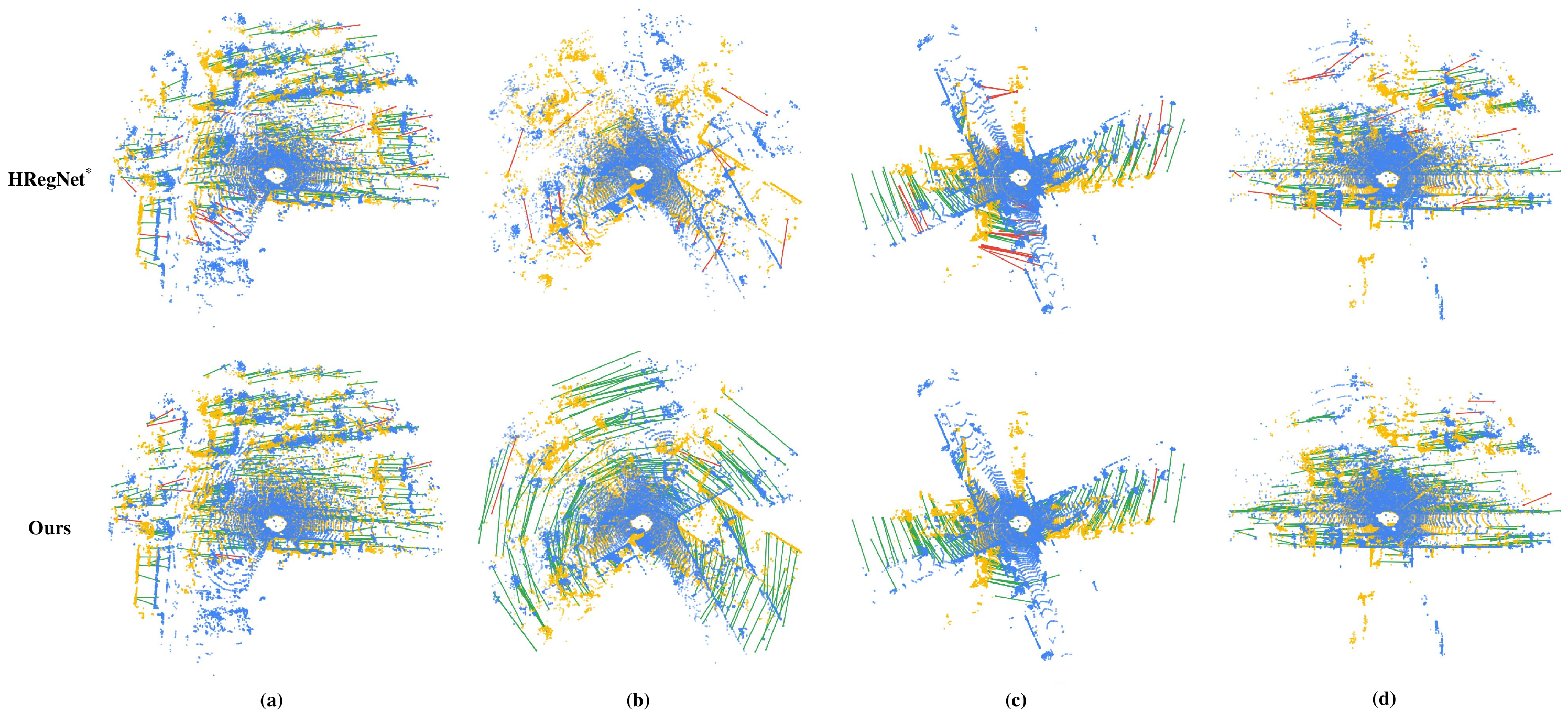}
\vspace{-3.2mm}
\caption{\textbf{Visualization of point cloud registration on KITTI.} Yellow and blue points indicate input point pair frames, with green lines showing accurate correspondences (inliers) and red lines showing inaccurate ones (outliers). This figure is best viewed in zoomed-in color mode. Compared to HRegNet$^{*}$ \cite{lu2023hregnet}, our method MemReg estimates more accurate correspondences.}
\vspace{-5mm}
\label{fig:visual}
\end{figure*}

\noindent\textbf{Comparison Results on nuScenes.} As in Table \ref{tab:reg2}, our method achieves a 100\% registration recall and outperforms previous optimization-based and learning-based methods with smaller registration errors, pushing the limit of performance further. Compared to the recent SOTA method HRegNet$^{*}$ \cite{lu2023hregnet}, our model results in only half of its translation errors (0.069\,m) and also a much lower rotation error (0.261\,deg).

\noindent\textbf{Comparison Results on Apollo-Southbay.} As in Table \ref{tab:reg3}, our method achieves 100\% registration recall and significantly surpasses all previous methods, especially in the RTE metric, establishing a new SOTA in registration performance.

\noindent\textbf{Comparison with Odometry or SLAM systems.} We also compare with SLAM methods in Table \ref{tab:reg4}. Our method not only surpasses standard odometry methods with sequential inputs like PWCLO \cite{wang2021pwclo}, DVLO \cite{liu2024dvlo}, and LAGLO \cite{tang2025enhanced}, but also outperforms methods that explicitly model multi-frame correlation like DELO \cite{ali2023delo}, DSLO \cite{zhang2024dslo}, and DVLO4D \cite{liu2025dvlo4d}. Our registration-only method also ha better registration results than LOAM \cite{zhang2014loam}, a standard LiDAR SLAM system with loop closure.

\noindent\textbf{Results Under Stricter Thresholds.} Since the threshold setting in \cite{lu2023hregnet} is rather loose, registration recalls are relatively saturated in these three datasets and the performance comparison with previous methods is unclear. Therefore, we vary a series of stricter thresholds in Fig. \ref{fig:thre} for KITTI and nuScenes datasets, where our method consistently outperforms previous optimization- and learning-based methods.

\subsection{Qualitative Results}
\noindent\textbf{Registration Samples.} As shown in Fig. \ref{fig:visual}, compared to HRegNet$^{*}$ \cite{lu2023hregnet}, introducing temporal 
\begin{wrapfigure}{r}{0.6\textwidth}
  \centering
\vspace{-3mm}
\includegraphics[width=1.0\linewidth]{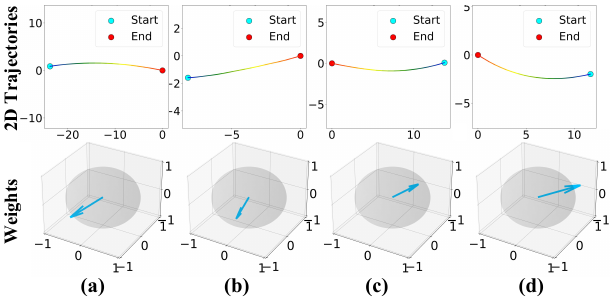}
\vspace{-7mm}
\caption{\textbf{Visualization of dynamic history weighting.} }
\vspace{-5mm}
\label{fig:visual_dynamic}
\end{wrapfigure}
information with memory buffers can significantly improve the ratio of successfully correlated keypoints. This demonstrates the effectiveness of our designed streaming temporal information on filtering outliers. 


\noindent\textbf{Weights in Dynamic History Weighting.} 
We display the weights $W_{2}$ and corresponding 2D trajectories in Fig. \ref{fig:visual_dynamic}, where the $W_{2}$ matrix is projected to a normalized 3D vector via PCA. Similar weights (lower figures) are observed in similar temporal trajectories (upper figures) such as (a)\&(b), (c)\&(d), demonstrating that our Dynamic History Weighting module can effectively learn weights from history trajectories.

\subsection{Generalization to Indoor Registration}

To demonstrate the generalization ability of our method, we also evaluate our method MemReg on indoor-level registration datasets: 3DMatch, 3DLoMatch \cite{zeng20173dmatch}, and ScanNet \cite{dai2017scannet}.
\begin{wrapfigure}{r}{0.56\textwidth}
\centering
\includegraphics[width=1\linewidth]{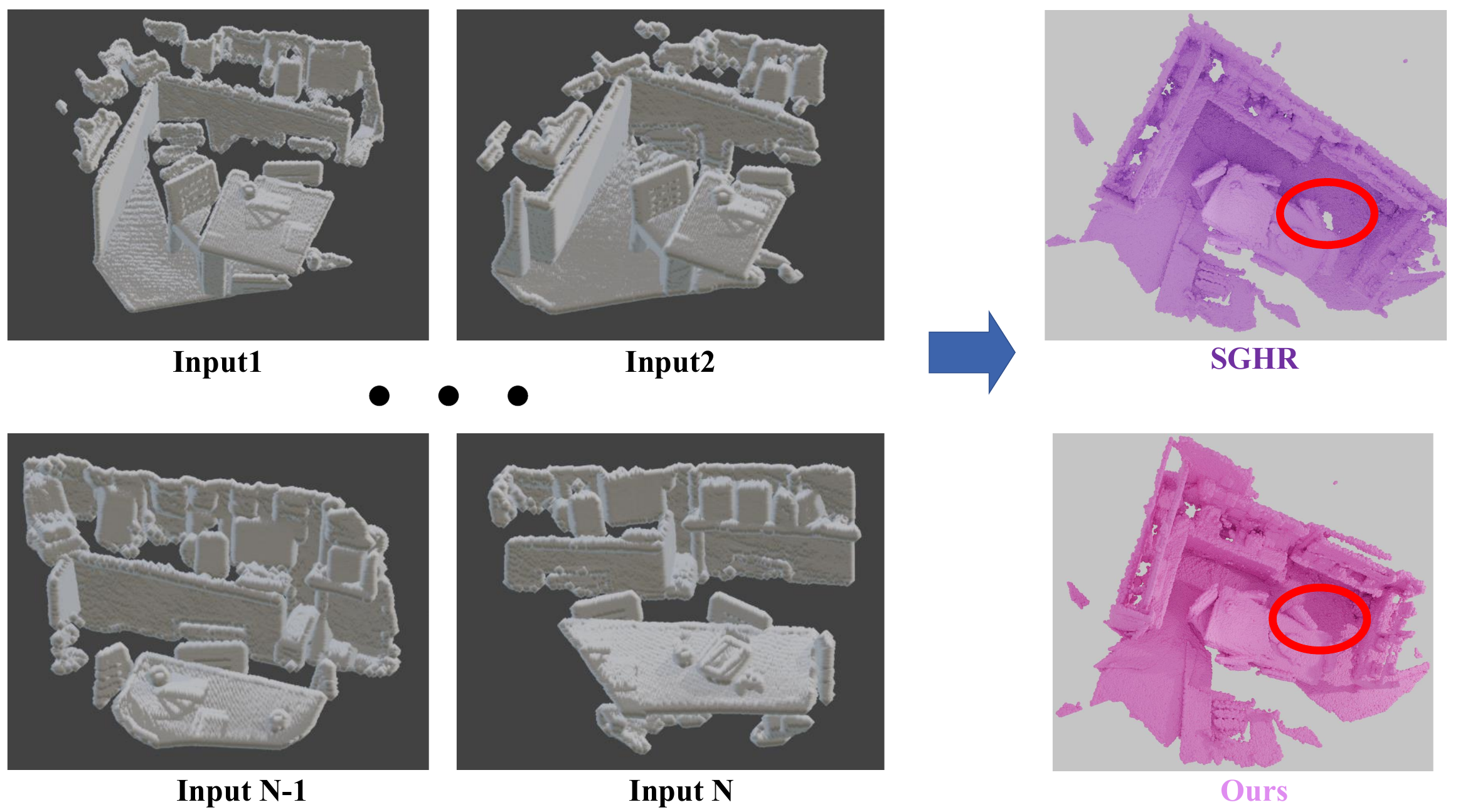}
\vspace{-7mm}
\caption{\textbf{Multiview indoor registration on 3DMatch \cite{zeng20173dmatch}.} We adapt our model to indoor scenes \cite{wang2023robust}, where our MemReg has more accurate pose estimation for areas with the red circle than the baseline method SGHR. }
\vspace{-2mm}
\label{fig:indoor}
\end{wrapfigure}
Current indoor methods can be divided into two categories in terms of different input frames: pairwise registration and multiview registration. 
As in Table \ref{tab:3dmatch_3dlomatch}, we first adapt our memory buffer module to general pairwise settings based on a recent SOTA method, HybridReg \cite{du2025hybridreg}. Compared to HybridReg, there is an obvious increase in registration recall and lower RRE, RTE on both 3DMatch and 3DLoMatch benchmarks. We also evaluate our method's compatibility in Table \ref{tab:indoor} on indoor multiview registration, where the network inputs are a series of partial indoor segments with overlap. We choose SGHR \cite{wang2023robust} as the baseline enhanced by our memory. Our method outperforms recent state-of-the-art methods \cite{arrigoni2016spectral,chatterjee2017robust,yew2021learning,lee2022hara,fang2024robust}, which are specially designed for the multiview indoor registration task. We also visualize the comparison in Fig. \ref{fig:indoor}. Our method registers input scans better compared to SGHR \cite{wang2023robust}, demonstrating our method's strong generalization ability for indoor-level scenes.

\setlength{\tabcolsep}{0.8mm}
\begin{table}[t]
\begin{minipage}{0.485\textwidth}
   \centering
\caption{\textbf{Generalization to pair-wise indoor registration on 3DMatch and 3DLoMatch.} We introduce our memory buffers into a recent SOTA indoor baseline HybridReg \cite{du2025hybridreg} for evaluation.}
  \label{tab:3dmatch_3dlomatch}
  \vspace{-6pt}
   \resizebox{1\columnwidth}{!}{
  \begin{tabular}{l|ccc|ccc}
    \hline \toprule
    \multirow{2}{*}{Methods} &
    \multicolumn{3}{c|}{3DMatch} &
    \multicolumn{3}{c}{3DLoMatch} \\
    \cline{2-4}\cline{5-7}\noalign{\smallskip}
    & RRE$\downarrow$ & RTE$\downarrow$ & RR$\uparrow$
    & RRE$\downarrow$ & RTE$\downarrow$ & RR$\uparrow$ \\
    \hline\noalign{\smallskip}
    FCGF \cite{choy2019fully} & 2.0 &6.6&85.1&3.2&10.0&40.1\\
    D3Feat \cite{bai2020d3feat}& 2.2 &6.7 & 81.6 & 3.4 &10.3&37.3\\
    PREDATOR \cite{huang2021predator} & 2.0& 6.4 & 89.0& 3.1&9.3&59.8\\
    CoFiNet \cite{yu2021cofinet} & 2.0 & 6.2 & 89.3&3.3&9.4&67.5\\
     RegTR    \cite{yew2022regtr}    & \textbf{1.6} & 4.9 & 92.0 & 2.8 & 7.7 & 64.8 \\
    GeoTransformer \cite{qin2022geometric}     & \textbf{1.6} & 5.3 & 91.5 & {2.5} & \textbf{7.4} & \bf{74.0} \\
    \hline\noalign{\smallskip}
    HybridReg \cite{du2025hybridreg} & \textbf{1.6} & 5.2 & 92.7 & 2.5 & 7.5 & 73.1 \\
      
    HybridReg + Ours & \textbf{1.6} & \textbf{4.8} & \textbf{92.9} & \textbf{2.4} & \textbf{7.4} & 73.5 \\  
    \bottomrule\hline
  \end{tabular}}
\end{minipage}
\hspace{3pt}
\begin{minipage}{0.485\textwidth}
   \centering
 \caption{\textbf{Generalization to multiview indoor registration on 3D(Lo)Match and ScanNet.} `Rot.' and `Trans.' are rotation and translation mean errors.}
  \vspace{-6pt}
  \label{tab:indoor}
  \setlength{\tabcolsep}{0.45mm}
   \resizebox{1\columnwidth}{!}
            {
  \begin{tabular}{l|cccc}
    \hline\toprule
    \multirow{2}[0]{*}{Methods} & 3DMatch & 3DLoMatch & \multicolumn{2}{c}{ScanNet}  \\
    \cline{4-5}
     & RR(\%)$\uparrow$ & RR(\%)$\uparrow$ & Rot. (\degree)$\downarrow$ & Trans. (m)$\downarrow$ \\
    \midrule
    EIGSE3 \cite{arrigoni2016spectral} & 40.1 & 26.5 & 40.6/37.1 & 0.88/0.84 \\
    L1-IRLS \cite{chatterjee2017robust} & 68.6 & 49.0 & 41.8/34.0 & 1.05/1.01 \\
    RotAvg \cite{chatterjee2017robust} & 77.2 & 63.0 & 38.5/31.6 & 0.96/0.83 \\
    LITS \cite{yew2021learning} & 80.8 & 65.2 & 24.9/19.9 & 0.65/0.56 \\
    HARA \cite{lee2022hara} & 83.8 & 79.1 & 34.7/31.3 & 0.86/0.17 \\
    \midrule
    SGHR \cite{wang2023robust} & 96.2 & 81.6 & 21.7/19.0 & 0.56/0.49 \\
    SMVR \cite{fang2024robust} & 96.2 & 82.0 & 19.8/17.5 & 0.55/0.45 \\
    SGHR + Ours & \bf97.1 & \bf82.8 & \bf19.2/16.3 & \bf0.52/0.44 \\
    \bottomrule\hline
  \end{tabular}}
\end{minipage}%
\vspace{-4mm}
\end{table}

\subsection{Ablation Studies}
In this section, we conduct comprehensive ablation studies to demonstrate the effectiveness of each proposed novelty.

\setlength{\tabcolsep}{2mm}
\begin{wraptable}{r}{0.56\textwidth}
   \centering
   \vspace{-6mm}
  \caption{\textbf{Ablation studies of the Dynamic History Weighting (DHW) on KITTI.}}
  \label{tab:ablation16}
  \scriptsize   
  \resizebox{1.00\linewidth}{!}
		{
  \begin{tabular}{l|cccc}
    \hline\toprule
    Model & RTE (m) & RRE (deg)  & Time (ms) \\
    \midrule
    w/o DHW & $0.038\pm 0.038$ & $0.136 \pm 0.113$  & 138 \\
    with estimator in \cite{liu2024dvlo} & $0.036\pm 0.040$ & $0.132 \pm 0.104$  & 135 \\
    with DHW & $\mathbf{0.032\pm0.036}$ & $\mathbf{0.122\pm 0.096}$  & \bf133 \\
    \bottomrule\hline
  \end{tabular}}
    \vspace{-5mm}
\end{wraptable}
\noindent\textbf{Dynamic History Weighting (DHW).} 
As in Table \ref{tab:ablation16}, removing DHW results in an 18.8\% higher RTE and 11.5\% higher RRE, indicating that DHW can effectively unify the history and current pose-related features. We also compare with another pose estimator in \cite{liu2024dvlo}, which has much higher RTE and RRE.

\noindent\textbf{Significance of Temporal Memory Buffers.} Table \ref{tab:ablation3} shows ablation studies for two memory buffers. Without the Memory Pose Buffer (MPB), the average RTE increases by 31.3\% and the average RRE by 27.9\%. Removing the Memory Feature Buffer (MFB) results in a 9.4\% higher average RTE and a 3.3\% higher average RRE.

\noindent\textbf{Different Temporal Lengths.} Table \ref{tab:ablation4} shows the comparison of different settings on temporal lengths in memory buffers. Accuracy increases largely from $T$=5 to 20 but is saturated when $T$$>$20. This is because 20 frames have enough historical observations for memory read-out. Points farther than 20 frames share less geometric and motion similarity, which may even degrade accuracy. More ablation studies about temporal interaction methods, rotation and translation thresholds are in the Appendix.

\section{Discussions}
\label{sec:diss}
Temporal information introduced by our designed memory buffers can address challenging cases where pair-wise registration works struggle, including low-overlap, dynamics, and occlusions.

\setlength{\tabcolsep}{0.2mm}
\begin{table}[t]
\begin{minipage}{0.485\textwidth}
   \centering
 \caption{\textbf{Ablation studies of the two memory buffers on KITTI.}}
    \vspace{-6pt}
	\label{tab:ablation3}
	\footnotesize
 \resizebox{1.00\columnwidth}{!}
		{
	\begin{tabular}{cc|cccc}
		\hline\toprule
		MFB&MPB & RTE (m) & RRE (deg)  & Time (ms) \\
		\midrule
		\xmark&\xmark &  $0.047\pm 0.046$ & $0.164 \pm 0.135$  & \bf126 \\
		\cmark& \xmark& $0.042\pm 0.042$ & $0.156\pm 0.124$  & 130 \\
		\xmark& \cmark& $0.035\pm 0.039$ & $0.126\pm 0.106$   & 132 \\
        \cmark & \cmark& $\mathbf{0.032\pm0.036}$ & $\mathbf{0.122\pm 0.096}$  & 133 \\
		\bottomrule\hline
	\end{tabular}}
\end{minipage}
\hspace{3pt}
\setlength{\tabcolsep}{1.5mm}
\begin{minipage}{0.485\textwidth}
   \centering
\caption{\textbf{Ablation studies of different temporal lengths on KITTI.}}
	\label{tab:ablation4}
	\footnotesize
    \vspace{-6pt}
    \resizebox{1.00\columnwidth}{!}
		{
	\begin{tabular}{c|cccc}
		\hline\toprule
		$T$ & RTE (m) & RRE (deg) & Time (ms) \\
		\midrule
		5 & $0.037\pm 0.041$ & $0.136\pm 0.116$   & \bf 128\\
		10 & $0.033\pm 0.036$ & $0.127\pm 0.101$   & 129 \\
       20 & $\mathbf{0.032\pm0.036}$ & $\mathbf{0.122\pm 0.096}$   & 133\\
        30 & $\mathbf{0.032}\pm 0.038$ & $0.129\pm 0.105$  & 140 \\
		\bottomrule\hline
	\end{tabular}}
\end{minipage}%
\vspace{-1mm}
\end{table}

\setlength{\tabcolsep}{1mm}
\begin{wraptable}{r}{0.56\textwidth}
   \centering
   \vspace{-7mm}
  \caption{\textbf{Evaluation on low-overlap input pairs with 20-frame and 30-frame intervals on KITTI. }}
 \vspace{2pt}
  \label{tab:robustness1}
  \scriptsize  
  \resizebox{1.00\linewidth}{!}
		{
  \begin{tabular}{l|c|ccc}
    \hline
    \toprule
    Method & Intervals & RTE (m) & RRE (deg) & Recall \\
    \midrule
    HRegNet* \cite{lu2023hregnet} & 20-frame & $0.085\pm0.152$ & $0.235\pm0.213$ & 96.2\% \\
    Ours & 20-frame & $\bf{0.041\pm0.039}$ & $\bf{0.148\pm0.127}$ & \textbf{100.0}\% \\
    \midrule
    HRegNet* \cite{lu2023hregnet} & 30-frame & $0.123\pm0.195$ & $0.315\pm0.302$ & 76.3\% \\
    Ours & 30-frame & $\bf{0.046\pm0.042}$ & $\bf{0.169\pm0.141}$ & \textbf{100.0}\% \\
    \bottomrule\hline
  \end{tabular}}
\vspace{-3mm}
\end{wraptable}

\begin{figure}[t]
\begin{minipage}{0.455\textwidth}
   \centering
\includegraphics[width=1.0\linewidth]{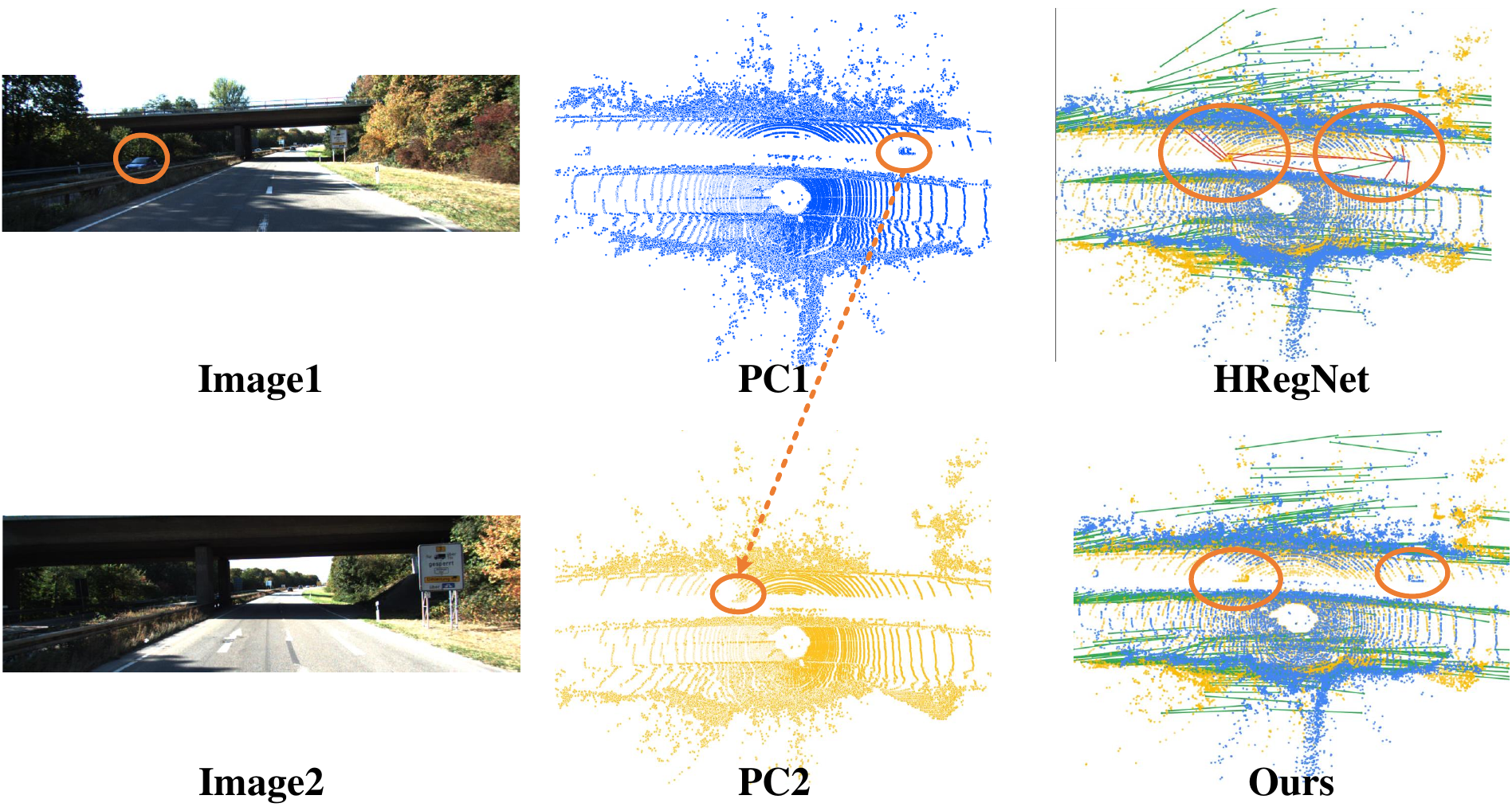}
\vspace{-6mm}
\caption{\textbf{Comparison on dynamics.} We evaluate on highway scenes with high dynamic cars (orange circled). HRegNet$^{*}$ \cite{lu2023hregnet} wrongly estimates correspondences on dynamics, while ours filters influence with no estimated correspondences on dynamics. }
\vspace{-2mm}
\label{fig:robustness2}
\end{minipage}
\hspace{1pt}
\begin{minipage}{0.515\textwidth}
   \centering
\includegraphics[width=1.0\linewidth]{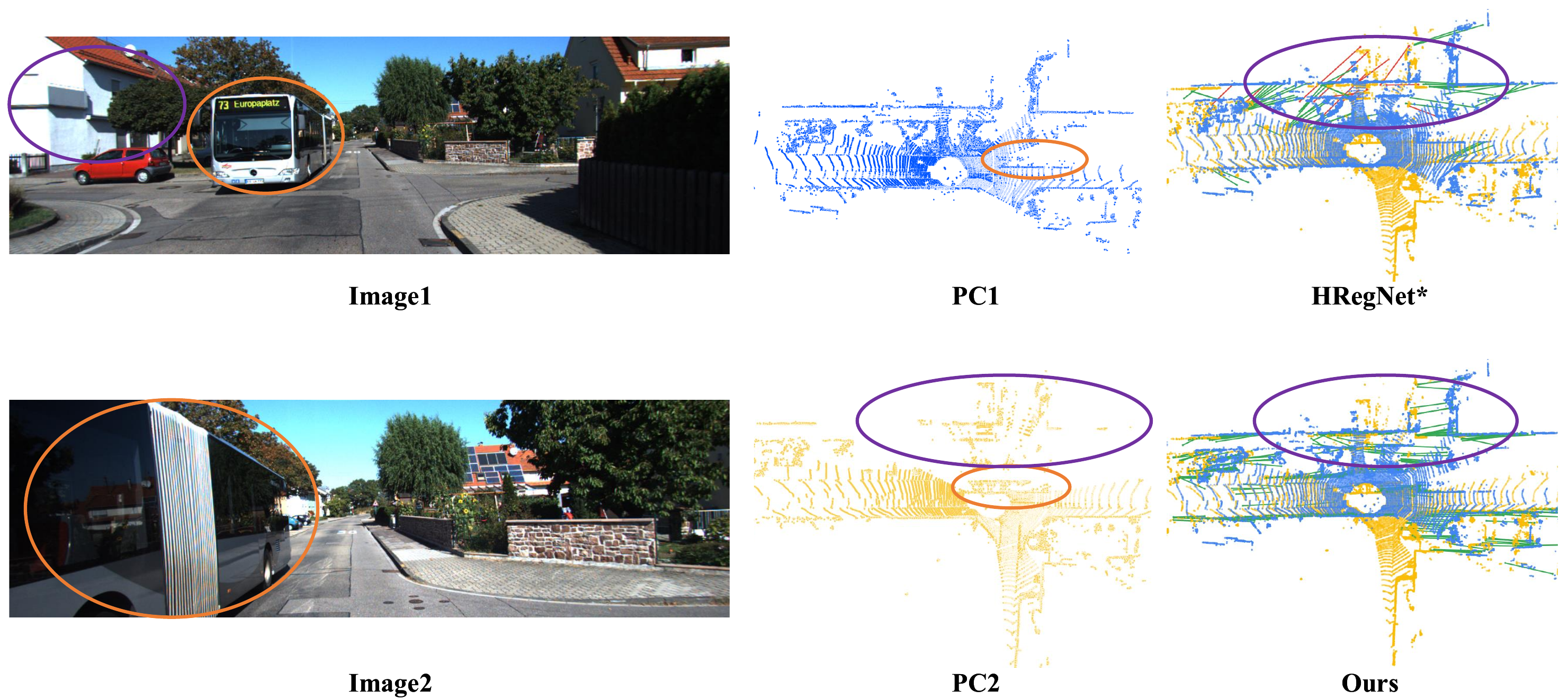}
 \vspace{-6mm}
\caption{\textbf{Comparison on scenes with significant occlusions.} For scenes with large occluded areas by the bus (orange circled). HRegNet$^{*}$ \cite{lu2023hregnet} wrongly estimates correspondences for occluded objects (purple circled), but ours can filter their influence with no correspondences estimated on occluded objects. }
\vspace{-4mm}
\label{fig:robustness3}
\end{minipage}
\end{figure}

\noindent\textbf{Low-Overlap Inputs.} Low-overlap registration is still an open issue because real-world LiDAR scans may have large distances or rotations, especially for low-frequency scanning LiDAR sensors, fast-moving cars, and street corners with large rotations. In these cases, the classical nearest neighboring strategy, e.g., KNN, may fail to capture precise point correspondences. In Table \ref{tab:robustness1}, we evaluate the low-overlap registration performance with larger frame intervals, which demonstrates the effectiveness of our designed memory buffers compared to the pair-wise baseline method HRegNet$^{*}$ \cite{lu2023hregnet}.

\noindent\textbf{Dynamics and Occlusions.} Dynamic or occluded objects introducing inconsistent motion patterns are commonly viewed as outliers in the outdoor registration task \cite{liu2023regformer}, which undermines the accuracy of consistent pose regression. We visualize the estimated correspondences for scenes with high dynamics in Fig. \ref{fig:robustness2} and significant occlusions in Fig. \ref{fig:robustness3}. Our method can mitigate their influence with fewer or even no estimated correspondences on these outlier points. Therefore, the subsequent pose regression would be less influenced by these outliers.

\section{Conclusion}

In this paper, we introduced a memory buffer using streaming history features to enhance outdoor point cloud registration. Two temporal buffers are designed to store feature-level and pose-level memories from previous timestamps, exploiting long-term history motion priors to address challenging outdoor scenes. Additionally, a Mamba-based temporal interaction module and a dynamic history weighting module are developed to merge historical and current features. Extensive experiments demonstrate the effectiveness of our proposed method on KITTI, nuScenes, and Apollo-Southbay datasets. Our MemReg also has excellent generalization ability on the indoor registration benchmarks like 3DMatch, 3DLoMatch, and ScanNet.

\section{Acknowledgments} 
This work was supported in part by the Natural Science Foundation of
China under Grant 62225309,U24A20278, 62361166632, the Explorers Program of Shanghai (Basic Research Funding) under Grant 25TS1415100, in part by the Shanghai Municipal Special Program for Basic Research on General AI Foundation Models under Grant 2025SHZDZX025G01.

{
\small
\bibliographystyle{unsrt}
\bibliography{main}
}


\appendix

\section{Overview}
\label{sec:overview}
The supplementary materials are structured as follows:
\begin{itemize}
\item We first provide more details about the network architecture in Sec. \ref{sec:arc}.
\item The evaluation metrics are described in Sec. \ref{sec:eva}.
\item Additional experiment results are presented in the Sec. \ref{sec:exp} to demonstrate the superiority of our method and the effectiveness of the proposed contributions.
\item More visualizations on the KITTI, nuScenes, and Apollo-Southbay datasets are provided in Sec. \ref{sec:vis}.
\item Additionally, we append the video demos of the outdoor registration performance on four sequences of KITTI (Seq.01, Seq.08, Seq.09, Seq.10) to the supplementary materials. 
\end{itemize}

\section{Network Architecture}
\label{sec:arc}

\subsection{Feature Extraction and Coarse Matching}
\label{sec:feature_extraction}
We follow HRegNet \citep{lu2023hregnet} to construct the hierarchical point feature extraction and coarse point matching layers.

\noindent\textbf{Feature Extraction.} Hierarchical point features are extracted as in \citep{lu2023hregnet}. For each layer, Weighted Farthest Point Sampling (WFPS) \citep{qi2017pointnet++} is first utilized to sample a set of center keypoints. Around each keypoint, K Nearest Neighbor (KNN) and Shared Multi-layer Perceptron (Shared MLP) are then used to search and aggregate the features of its k nearest neighboring points:
\begin{equation}
    \label{eq:feature gathering}
     {f_i} = \max_{k=1,2,\cdots,\text{K}} \text{MLP}(({x_i^k}-{x_i})\oplus{f_i^k}),
\end{equation}
where ${x_i^k}$ denotes the $k$-th nearest point and ${f_i^k}$ is the corresponding point feature. $\mathop{\text{Max}}$ is the Max-pooling operation. 

\noindent\textbf{Coarse Point Matching.} The core issue in point cloud registration is to find precise and reliable correspondences between two frames. At the coarsest layer, where points commonly possess the largest receptive field, finding accurate correspondences can guarantee the generally reliable registration performance. To achieve this, we consider geometric features, descriptor features, and similarity features from current point pairs as in HRegNet \citep{lu2023hregnet}. Specifically, for each keypoint in $PC_{\mathrm{t}}$ as the clustering center, we perform KNN to select a set of neighboring candidate points in $PC_{\mathrm{t+1}}$. The center keypoint and its K neighboring points form a cluster for feature embedding. Geometric features consist of both the absolute coordinates and the relative distance of the keypoint and its neighboring points. Descriptor features include the saliency uncertainties of keypoints and also the local descriptors for each keypoint and its neighboring points. Similarity features are considered in both bilateral consensus and neighborhood consensus. Bilateral consensus ensures the bi-directional optimal matching scores in both the forward (keypoints in $PC_{\mathrm{t}}$ searching KNN in $PC_{\mathrm{t+1}}$) and the reverse (keypoints in $PC_{\mathrm{t+1}}$ searching KNN in $PC_{\mathrm{t}}$) processes. Neighborhood consensus indicates that the neighboring keypoints around the correspondence keypoints should possess similar features. The outputs of the coarse matching layer are cross-frame features $F_{t}$ and initially estimated pose $R_{t}$, $t_{t}$. However, these features and poses contain no temporal information, which will interact with and update temporal memory buffers (Secs. 3.3-3.5 in the main paper). 

\subsection{Mamba Architecture}
\begin{figure}[t]
\centering
\includegraphics[trim=3pt 0pt 3pt 0pt, clip=true, width=0.7\linewidth]{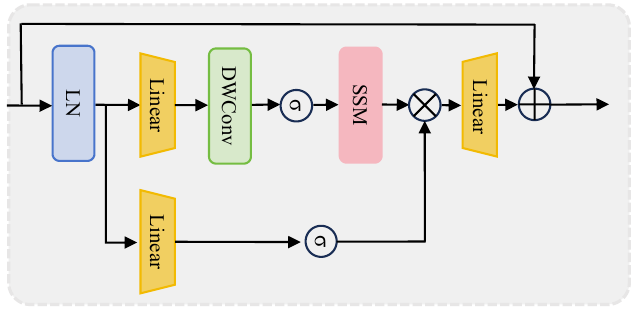}
\vspace{-4mm}
\caption{\textbf{The architecture of the Mamba module.} We utilize the standard Mamba \citep{gu2023mamba} module without any task-specific design.}
\label{fig:mamba}
\end{figure}

Recently, a novel architecture named Mamba \citep{gu2023mamba} has shown promising effectiveness, especially for long-sequence modeling ability, outperforming Transformer-based counterparts in image processing \citep{liu2024vmamba}, video understanding \citep{li2024videomamba}, and point cloud \citep{liang2024pointmamba,liu2024point} fields. In this paper, we adopt the vanilla Mamba architecture \citep{gu2023mamba} in our temporal encoding layer as illustrated in Fig. \ref{fig:mamba}. Without any task-specific designs, our Mamba-based temporal encoding turns out to be effective enough to embed long-term dependencies for history and current motion features.

\setlength{\tabcolsep}{5mm}
\begin{table}[tb]
  \centering
  \caption{{\bf{Implementation details of coarse matching and pose refinement layers.}} In the coarse matching and pose refinement layers, `N' denotes the number of the selected neighboring points in KNN. `MLP1' and `MLP2' represent the output dimension of two MLP layers, respectively, for correspondence prediction.}
  \resizebox{1\linewidth}{!}{
    \begin{tabular}{l|ccccccccc}
    \hline\toprule
    \multicolumn{2}{c}{Module} & \multicolumn{2}{|c}{Layer} & \multicolumn{2}{c}{N} & \multicolumn{2}{c}{MLP1} & \multicolumn{2}{c}{MLP2} \\
    \midrule
    \multicolumn{2}{c}{{Coarse Matching}} & \multicolumn{2}{|c}{Layer 1} & \multicolumn{2}{c}{8} & \multicolumn{2}{c}{[128,\,128,\,128]} & \multicolumn{2}{c}{[128,\,128,\,1]} \\
    
     \midrule
    \multicolumn{2}{c}{Module} & \multicolumn{2}{|c}{Layer} & \multicolumn{2}{c}{N} & \multicolumn{2}{c}{MLP1} & \multicolumn{2}{c}{MLP2} \\
    \midrule
    \multicolumn{2}{c}{\multirow{2}[0]{*}{Pose Refinement}} &\multicolumn{2}{|c}{Layer 2}& \multicolumn{2}{c}{8} & \multicolumn{2}{c}{[256,\,256,\,256]} & \multicolumn{2}{c}{[256,\,256,\,1]} \\
    \multicolumn{2}{c}{} & \multicolumn{2}{|c}{Layer 3} & \multicolumn{2}{c}{8} & \multicolumn{2}{c}{[512,\,512,\,512]} & \multicolumn{2}{c}{[512,\,512,\,1]} \\
    
    \bottomrule\hline
    \end{tabular}%
  }
  \label{tab:feature pyramid}%
\end{table}%

\subsection{Pose Refinement Layer}
We adopt the same pose refinement process as in HRegNet \citep{lu2023hregnet}, where finer registration layers only consider the geometric features and descriptor features. Similarity features are abandoned due to high computational costs. With the correlated features, a soft correspondence layer is introduced to predict keypoint correspondences and their confidence scores \citep{lu2021hregnet}. Finally, the Weighted Kabsch algorithm is applied to calculate the transformation matrices $\Delta R$ and $\Delta t$. Without loss of generality, taking layer $l$ as an example, the refinement layer can be represented by:
\begin{equation}
    \label{eq:ref}
     R^{l} = \Delta R \times R^{l+1},
\end{equation}
\begin{equation}
    \label{eq:ref2}
     t^{l} = \Delta R \times t^{l+1} + \Delta t,
\end{equation}
where $R^{l}, t^{l}$ indicate the refined poses at a finer layer. $R^{l+1}, t^{l+1}$ indicate the un-refined poses at a coarser layer. More implementation details are provided in Table \ref{tab:feature pyramid}.

\section{Evaluation Metrics}
\label{sec:eva}
Here, we provide the equations to calculate RTE (Relative Translation Error) and RRE (Relative Rotation Error). RTE is calculated by:
\begin{equation}
    \label{eq:metric}
     \text{RTE} = || \hat{t} -t ||_{2}.
\end{equation}
RRE is defined as:
\begin{equation}
    \label{eq:metric2}
     \text{RRE} = \arccos(\text{Tr}(\hat{R}^\text{T}R - 1)/2),
\end{equation}
where $\hat{R}, \hat{t}$ indicate the respective estimated rotation matrix and translation vector, and $R, t$ are the ground truth rotation matrix and translation vector.

\setlength{\tabcolsep}{12mm}
\begin{table}
	\centering
	\caption{\textbf{More comparison results on the KITTI odometry \citep{geiger2012we}.} }
	\label{tab:kittireg}
	\footnotesize
    \resizebox{1.00\columnwidth}{!}
            {
	\begin{tabular}{l|ccc}
       
		\toprule
		\multirow{2}{*}{Methods}&\multicolumn{3}{|c}{KITTI Odometry}\\
		\cmidrule(r){2-4}
		~ & RTE (m) & RRE (deg) & Recall  \\
		\midrule
       
        PosDiffNet \citep{she2024posdiffnet}& $0.066 $  &$0.240 $  & 99.80\% \\
        TrT-Net \citep{chen2024fast} & $0.063 $  & $0.230 $  & 99.80\% \\
        
        ML-SemReg \citep{yan2024ml} & $0.052$  &$0.200$ & 97.91\% \\
        

        RAP \citep{pan2026register}  & $0.043$  &$0.139$ & \bf100.0\% \\
        
        \midrule
        Ours & $\bf{0.032}$ & $\bf{0.122}$ & \textbf{100.0}\% \\
		\bottomrule
	\end{tabular}
 }
\end{table}

\section{Additional Experimental Results}
\label{sec:exp}

\textbf{More Comparison Results.} We also supplement more comparisons with recent state-of-the-art methods including PosDiffNet \citep{she2024posdiffnet}, TrT-Net \citep{chen2024fast}, ML-SemReg \citep{yan2024ml}, and RAP \cite{pan2026register}. Since they only release the average rotation and translation metrics, we also follow them for a fair comparison. As shown in Table \ref{tab:kittireg}, our method has substantial performance advantages compared to these methods, which demonstrates the superiority of our proposed temporal buffers.

\vspace{3pt}\noindent\textbf{Performance Under More Challenging Thresholds.} We also compare with baseline HRegNet$^{*}$ \citep{lu2023hregnet} in terms of different settings $\theta_{T}$ and $\theta_{R}$ in Table \ref{tab:ablation1} and Table \ref{tab:ablation2}. From these tables, much obviously higher recalls, lower RRE, and lower RTE can be observed compared to HRegNet$^{*}$. This demonstrates the great potential of our method in more challenging outdoor scenes with larger rotations and translations.

\vspace{3pt}\noindent\textbf{Ablation Studies for Various Temporal Interaction Methods.} For effective memory read-out, we explore different temporal interaction methods, including temporal averaging, GRU-based, Attention-based, and Mamba-based approaches. As shown in Table \ref{tab:ablation5}, the Mamba-based method outperforms the others, particularly in reducing rotation errors. This improvement is due to the effectiveness of the Mamba model in capturing long-term temporal dependencies.

\vspace{3pt}\noindent\textbf{Ablation Studies for Different Mamba Layers.} We also vary the number of Mamba layers from 1 to 3 in Table \ref{tab:ablation6}. Performance remains stable across all settings, demonstrating our model is insensitive to the specific design choice in Mamba. Notably, we choose 1 as the layer number of the Mamba module because increasing the number cannot bring a significant performance increase.

\vspace{3pt}\noindent\textbf{Ablation Studies for Learnable Parameters in DHW.} Learnable weights are of great significance to the dynamic fusion of history and current pose-related features in the Dynamic History Weighting (DHW). To prove this, we replace the learnable weighting matrices $W_{1}, W_{2}$ with fixed ones. As shown in Table \ref{tab:ablation7}, using fixed \(W_1\) or \(W_2\) increases both RTE and RRE compared to the learnable DHW, showing that dynamic weighting truly works.

\begin{table}
	\centering
	\caption{\textbf{Comparison results with HRegNet$^{*}$ \citep{lu2023hregnet} under different translation thresholds $\theta_{T}$ on the KITTI dataset.} $\theta_{R}$ is set to 1deg.}
	\label{tab:ablation1}
	\footnotesize
    \resizebox{1.00\columnwidth}{!}
		{
	\begin{tabular}{l|l|ccc}
		\toprule
		$\theta_{T}$ (m) & Method & RTE (m) & RRE (deg) & Recall  \\
		\midrule
		\multirow{2}{*}{0.05} &HRegNet$^{*}$& $0.0302\pm0.0110$ & $0.1583\pm 0.1155$ & 72.08\%  \\
		\cmidrule(r){2-5}
    		&Ours & $\bf{0.0243\pm 0.0109}$ & $\bf{0.1175 \pm 0.0837}$ & \bf87.50\%  \\
		\midrule
        \multirow{2}{*}{0.1} &HRegNet$^{*}$& $0.0406\pm0.0201$ & $0.1606\pm 0.1195$ & 95.60\%  \\
		\cmidrule(r){2-5}
    		&Ours & $\bf{0.0294\pm 0.0172}$ & $\bf{0.1184 \pm 0.0889}$ & \bf97.93\%  \\
        \midrule
        \multirow{2}{*}{0.2} &HRegNet$^{*}$& $0.0444\pm0.0271$ & $0.1613\pm 0.1192$ & 97.62\%  \\
		\cmidrule(r){2-5}
    		&Ours & $\bf{0.0310\pm 0.0225}$ & $\bf{0.1198 \pm 0.0881}$ & \bf98.13\%  \\
        \midrule
        \multirow{2}{*}{0.3} &HRegNet$^{*}$& $0.0455\pm0.0308$ & $0.1616\pm 0.1194$ & 99.52\%  \\
		\cmidrule(r){2-5}
    		&Ours & $\bf{0.0318\pm 0.0259}$ & $\bf{0.1197 \pm 0.0879}$ & \bf99.88\%  \\
		
		\bottomrule
	\end{tabular}}
\end{table}

\begin{table}
	\centering
	\caption{\textbf{Comparison results with HRegNet$^{*}$ \citep{lu2023hregnet} under different rotation thresholds $\theta_{R}$ on the KITTI dataset.} $\theta_{T}$ is set to 0.05m.}
	\label{tab:ablation2}
	\footnotesize
    \resizebox{1.00\columnwidth}{!}
		{
	\begin{tabular}{l|l|ccc}
		\toprule
		$\theta_{R}$ (deg) & Method & RTE (m) & RRE (deg) & Recall  \\
		\midrule
		\multirow{2}{*}{0.25} &HRegNet$^{*}$& $0.0302\pm0.0109$ & $0.1210\pm 0.0627$ & 57.20\%  \\
		\cmidrule(r){2-5}
    		&Ours & $\bf{0.0241\pm 0.0109}$ & $\bf{0.1006 \pm 0.0538}$ & \bf81.10\%  \\
		\midrule
        \multirow{2}{*}{0.5} &HRegNet$^{*}$& $0.0304\pm0.0109$ & $0.1524\pm 0.0973$ & 67.38\%  \\
		\cmidrule(r){2-5}
    		&Ours & $\bf{0.0242\pm 0.0109}$ & $\bf{0.1158 \pm 0.0781}$ & \bf87.12\%  \\
        \midrule
        \multirow{2}{*}{1} &HRegNet$^{*}$& $0.0304\pm0.0109$ & $0.1593\pm 0.1121$ & 68.41\%  \\
		\cmidrule(r){2-5}
    		&Ours & $\bf{0.0243\pm 0.0109}$ & $\bf{0.1192 \pm 0.0882}$ & \bf87.71\%  \\
        \midrule
        \multirow{2}{*}{2} &HRegNet$^{*}$& $0.0304\pm0.0109$ & $0.599\pm 0.1147$ & 68.45\%  \\
		\cmidrule(r){2-5}
    		&Ours & $\bf{0.0243\pm 0.0109}$ & $\bf{0.1193 \pm 0.0890}$ & \bf87.72\%  \\
		
		\bottomrule
	\end{tabular}}

\end{table}

\begin{table}
	\centering
	\caption{\textbf{Ablation studies of temporal encoding methods on KITTI.}}
	\label{tab:ablation5}
	\footnotesize
    \resizebox{1.00\columnwidth}{!}
		{
	\begin{tabular}{l|cccc}
		\toprule
		Model & RTE (m) & RRE (deg) & Recall &Time (ms) \\
		\midrule
        Temporal Averaging &$0.045\pm 0.041$ & $0.143\pm 0.117$ & \textbf{100\%} &131 \\
		GRU & $0.034\pm 0.039$ & $0.133\pm 0.115$ & \textbf{100\%} &135 \\
        Attention & $0.033\pm 0.039$ & $0.138\pm 0.121$ & \textbf{100\%} &136 \\
        Mamba (Ours) & $\mathbf{0.032\pm0.036}$ & $\mathbf{0.122\pm 0.096}$ & \textbf{100\% }&\bf133 \\
		\bottomrule
	\end{tabular}}
\end{table}

\setlength{\tabcolsep}{6mm}
\begin{table}[t]
  \caption{\textbf{Ablation studies for different Mamba layers on KITTI.}}
  \label{tab:ablation6}
  \centering
  \resizebox{1\columnwidth}{!}{
    \begin{tabular}{l|c|c|c}
      \hline
      Mamba layers & RTE (m)              & RRE (deg)             & Recall   \\
      \hline
      1               & $\mathbf{0.032\pm0.036}$ & $0.122\pm 0.096$ & \textbf{100\%}    \\
      2               & $\mathbf{0.032}\pm0.037$    & $\mathbf{0.120}\pm0.103$     & \textbf{100}\%    \\
      3               & $\mathbf{0.032}\pm0.037$ & $0.121\pm\mathbf{0.092}$ & \textbf{100\%} \\
      \hline
    \end{tabular}
  }
  \label{tab:ablation_mamba}
\end{table}

\setlength{\tabcolsep}{5mm}
\begin{table}[htbp]
  \centering
  \caption{\textbf{Ablation studies of the learnable parameters ($W_1$, $W_2$) in Dynamic History Weighting.}}
    
  \resizebox{1\columnwidth}{!}{
    \begin{tabular}{l|c|c|c}
      \hline
      \ Method & RTE (m)              & RRE (deg)             & Recall   \\
      \hline
     
      Ours w/ fixed $W_1$  & $0.036\pm0.039$    & $0.131\pm0.103$     & \textbf{100}\%    \\
      Ours w/ fixed $W_2$  & $0.037\pm0.041$    & $0.134\pm0.107$     & \textbf{100}\%    \\
       Ours               & $\mathbf{0.032\pm0.036}$ & $\mathbf{0.122\pm 0.096}$ & \textbf{100\%}    \\
      \hline
    \end{tabular}
  }
  \label{tab:ablation7}
\end{table}

\begin{figure*}[t]
\centering
\includegraphics[width=1.0\linewidth]{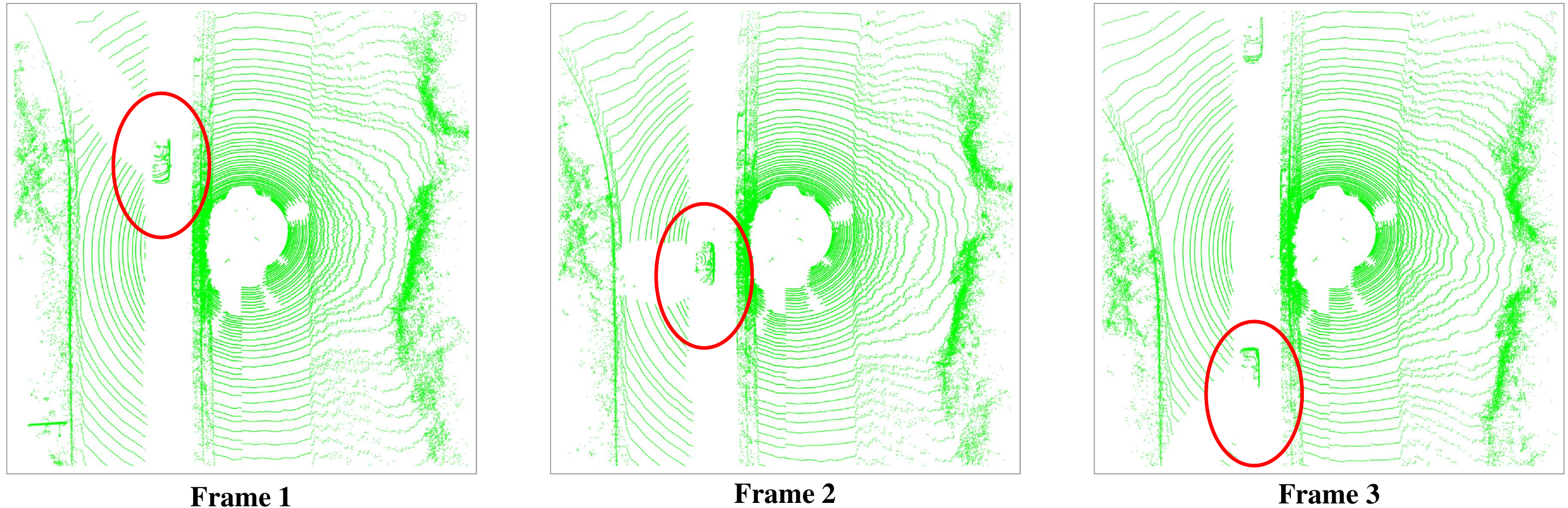}
\vspace{-6mm}
\caption{\textbf{A dynamic case in KITTI.} In a highway scene, partial observations hinder the recognition of the highly dynamic car.  }
\vspace{-2mm}
\label{fig:case}
\end{figure*}

\begin{figure}[t]
\centering
\includegraphics[width=0.8\linewidth]{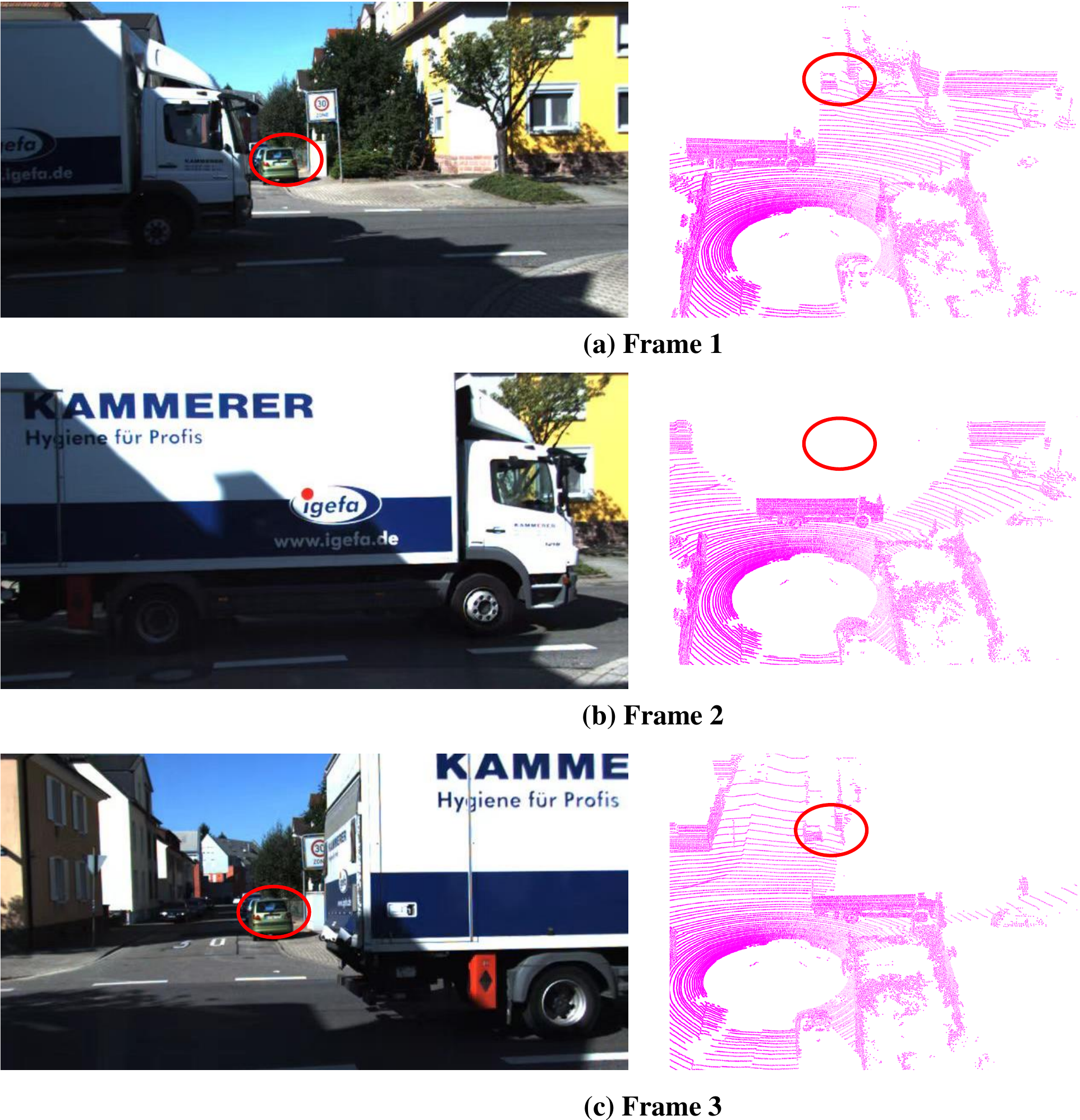}
\vspace{-4mm}
\caption{\textbf{An occlusion case in KITTI.} 
A pairwise pose regression can fail in occluded scenarios, e.g., the static car in the red circle in Frame 2, breaking the correlation in Frames 1\&2 and 2\&3. 
}
\vspace{-2mm}
\label{fig:case2}
\end{figure}

\begin{figure}[t]
\centering
\includegraphics[width=1.0\linewidth]{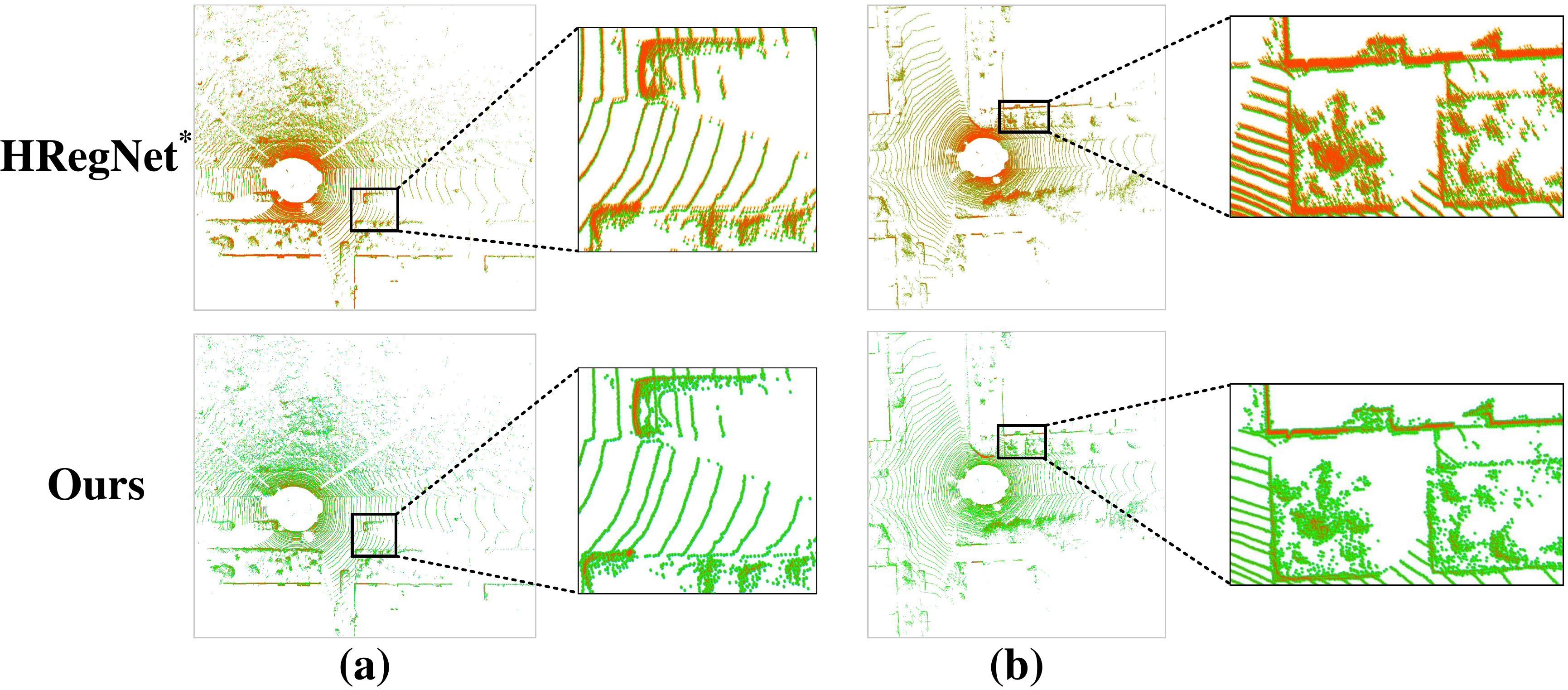}
\vspace{-6mm}
\caption{\textbf{Registration error visualization.} Our estimation (blue points) shows much smaller errors (red line) compared to HRegNet$^{*}$ \citep{lu2023hregnet} (yellow points) relative to GT (green points).}
\vspace{-4mm}
\label{fig:error}
\end{figure}

\section{Visualization Results}
\label{sec:vis}

\textbf{Outliers: Dynamic and Occlusion Cases.} Dynamics and occlusions are two main challenging outlier cases for outdoor point cloud registration \citep{liu2023regformer}. Here, in Fig. \ref{fig:case}, we first show one case of dynamics. As illustrated in the figure, when a dynamic car runs toward the ego-car at a high speed, parts of the dynamic car will be unobserved. For example, given only Frame 2 and Frame 3, the registration is difficult as the head of the car is not scanned in Frame 3. However, introducing more temporal inputs (Frame 1) will provide more geometric cues of the dynamic car and help recognize this outlier. Another common outlier is occlusion. For example, the occluded car in Frame 2 of Fig. \ref{fig:case2} is often seen as an outlier with pairwise inputs (e.g., Frames 1\&2), reducing registration accuracy. Sequential observations help identify the occlusion, as the car reappears in Frame 3 and can be correctly matched with Frame 1. 

\vspace{3pt}
\noindent\textbf{Registration Errors.} We display the registration errors of two scenes in Fig. \ref{fig:error}. Compared to HRegNet$^{*}$ \citep{lu2023hregnet}, our method generates much smaller errors thanks to the sufficient exploration of the temporal information.

\begin{figure*}[t]
\centering
\includegraphics[width=1.0\linewidth]{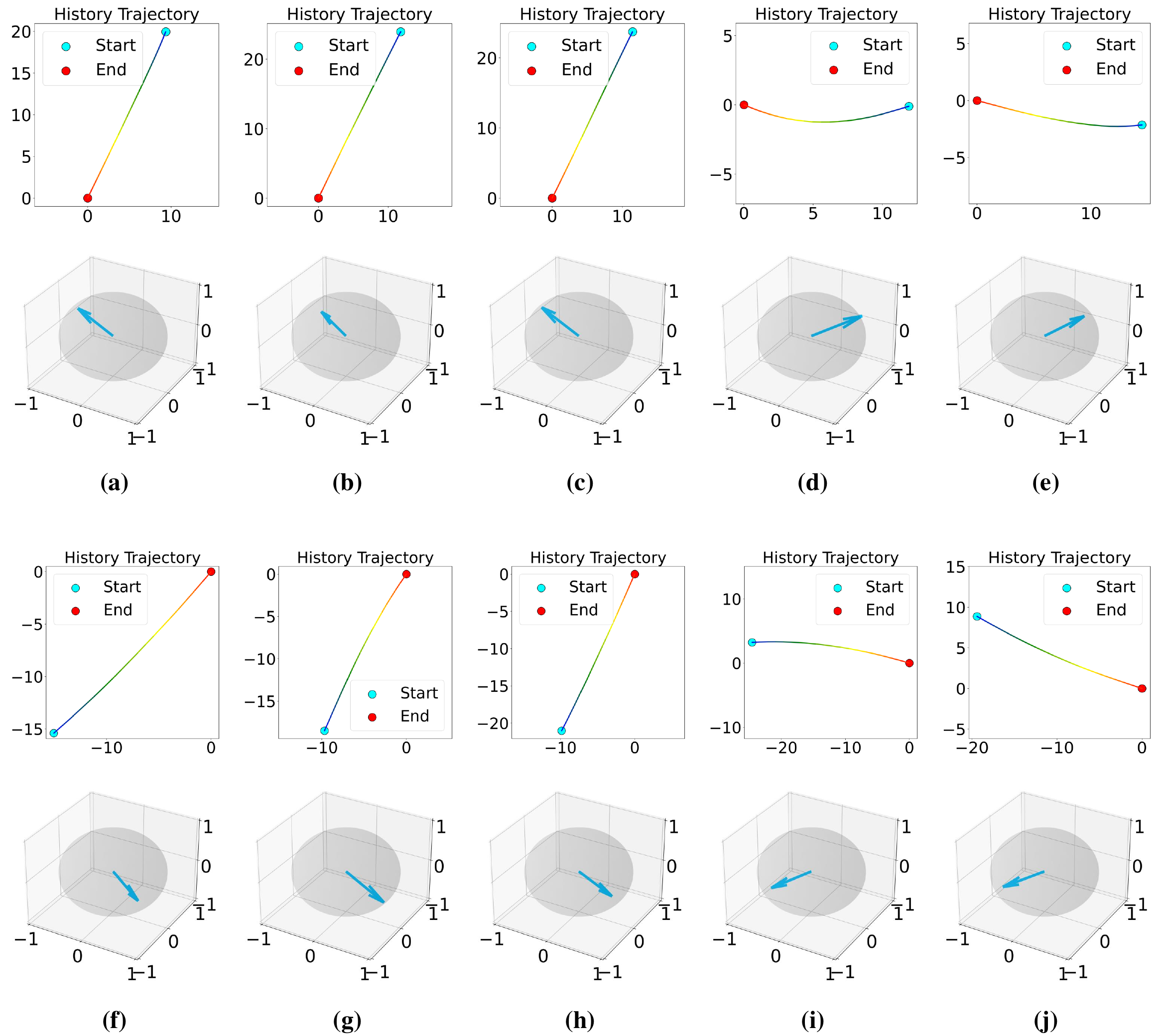}
\vspace{-3mm}
\caption{\textbf{The visualization of Dynamic History Weighting.} We show some samples about 2D trajectories (upper figures) and their corresponding weights (lower figures), which are generated from the PCA analysis of the weight matrix $W_{2}$.}
\vspace{-4mm}
\label{fig:dyn}
\end{figure*}

\vspace{3pt}
\noindent\textbf{Dynamic Weighting Visualization.} 
We visualize the weighting matrix using PCA and its corresponding 2D trajectories in Fig. \ref{fig:dyn}. From the figure, it is clear that similar historical trajectories often share similar weights, e.g., (a, b, c), (d, e), (f, g, h), and (i, j). This demonstrates the effectiveness of our Dynamic History Weighting method in learning weights from historical poses to influence the current motion features.

\begin{figure*}[t]
\centering
\includegraphics[trim=3pt 0pt 3pt 0pt, clip=true, width=1.0\linewidth]{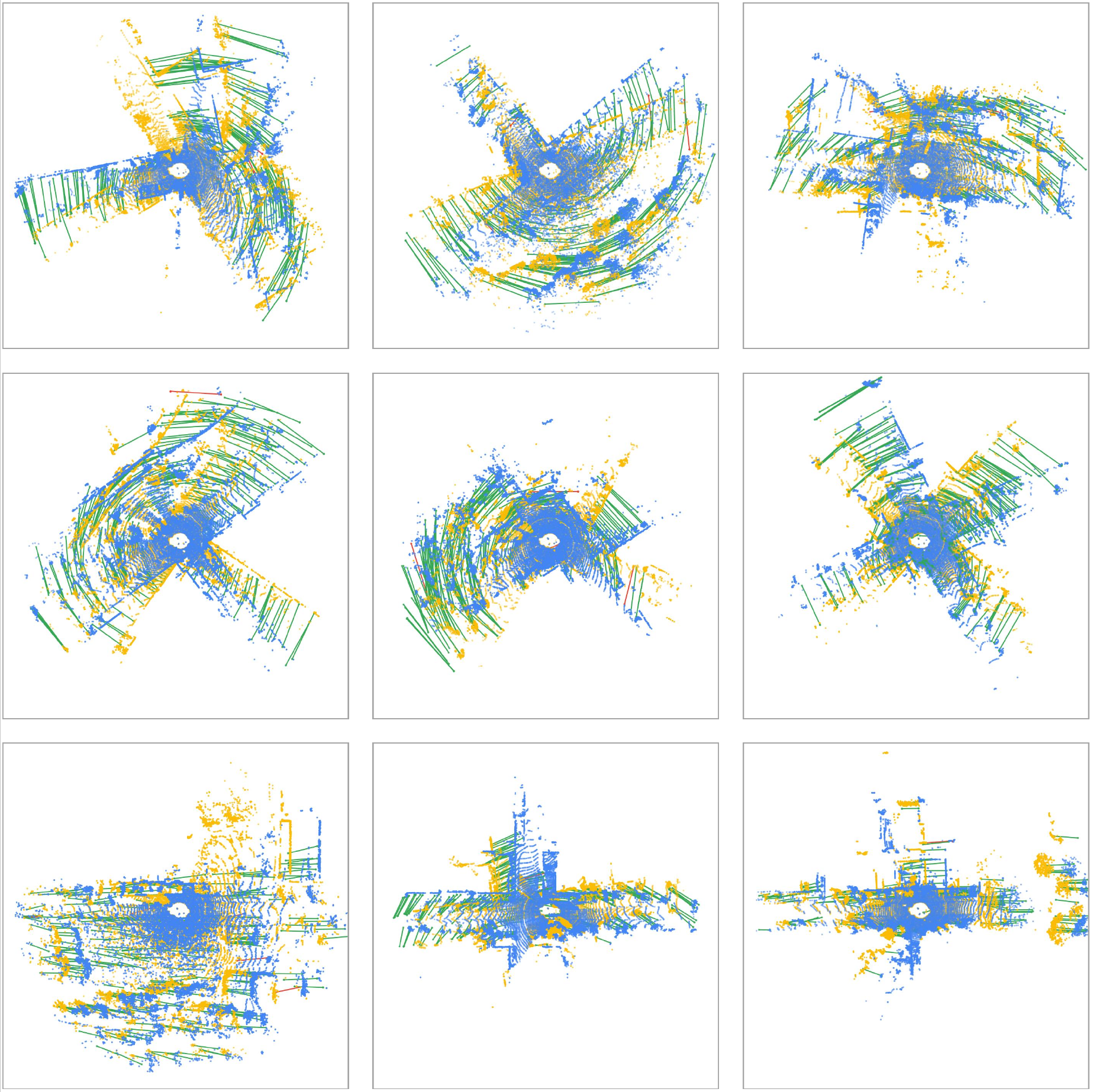}
\vspace{-4mm}
\caption{\textbf{The visualization of point cloud registration on KITTI.} Yellow and blue points indicate input point pair frames, with green lines showing accurate correspondences (inliers) and red lines showing inaccurate ones (outliers). This figure is best viewed in zoomed-in color mode.}
\vspace{-4mm}
\label{fig:visual1}
\end{figure*}

\begin{figure*}[t]
\centering
\includegraphics[trim=3pt 0pt 3pt 0pt, clip=true, width=1.0\linewidth]{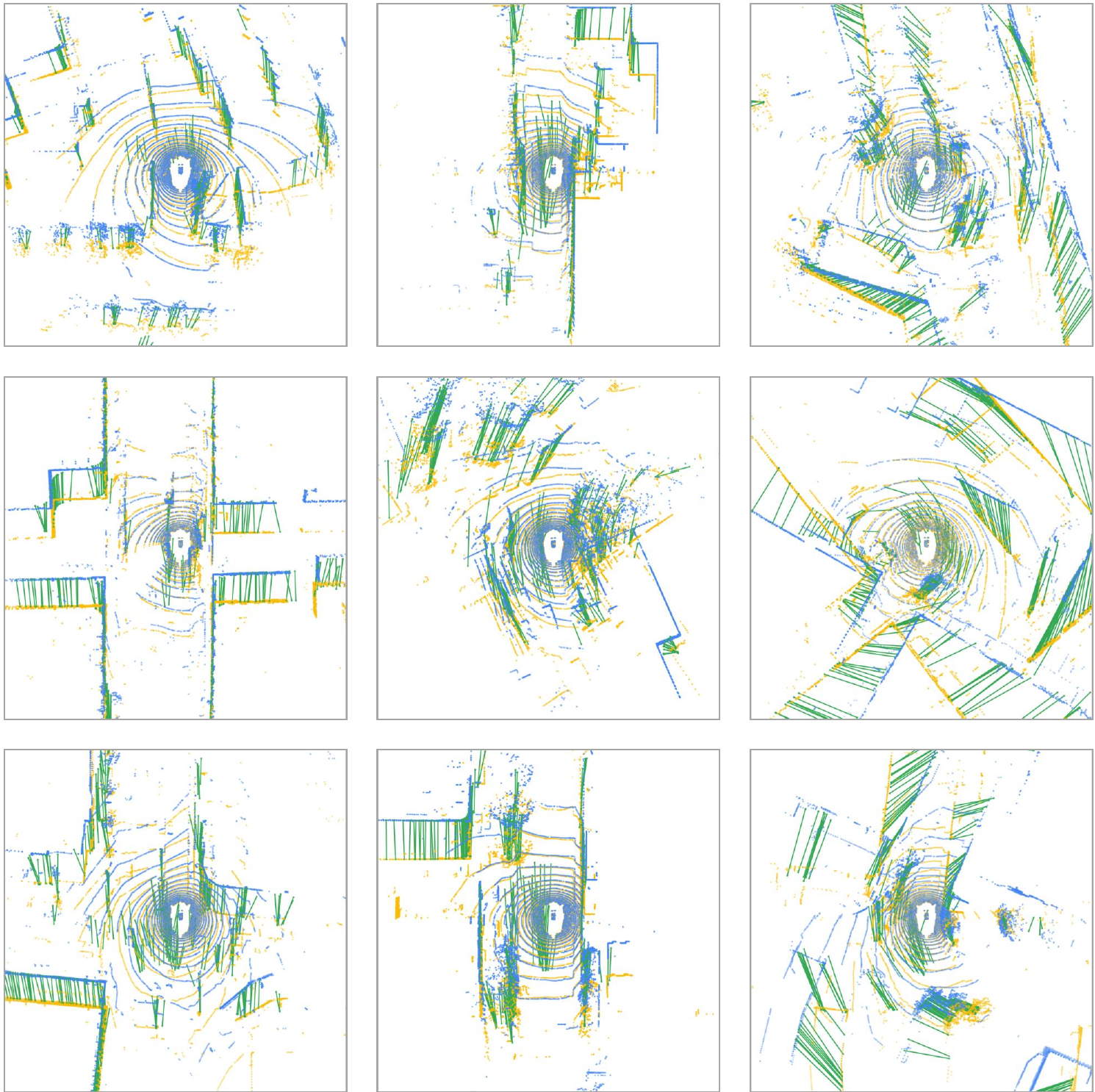}
\vspace{-4mm}
\caption{\textbf{The visualization of point cloud registration on nuScenes.} Yellow and blue points indicate input point pair frames, with green lines showing accurate correspondences (inliers) and red lines showing inaccurate ones (outliers). This figure is best viewed in zoomed-in color mode.}
\vspace{-4mm}
\label{fig:visual2}
\end{figure*}

\begin{figure*}[t]
\centering
\includegraphics[trim=3pt 0pt 3pt 0pt, clip=true, width=1.0\linewidth]{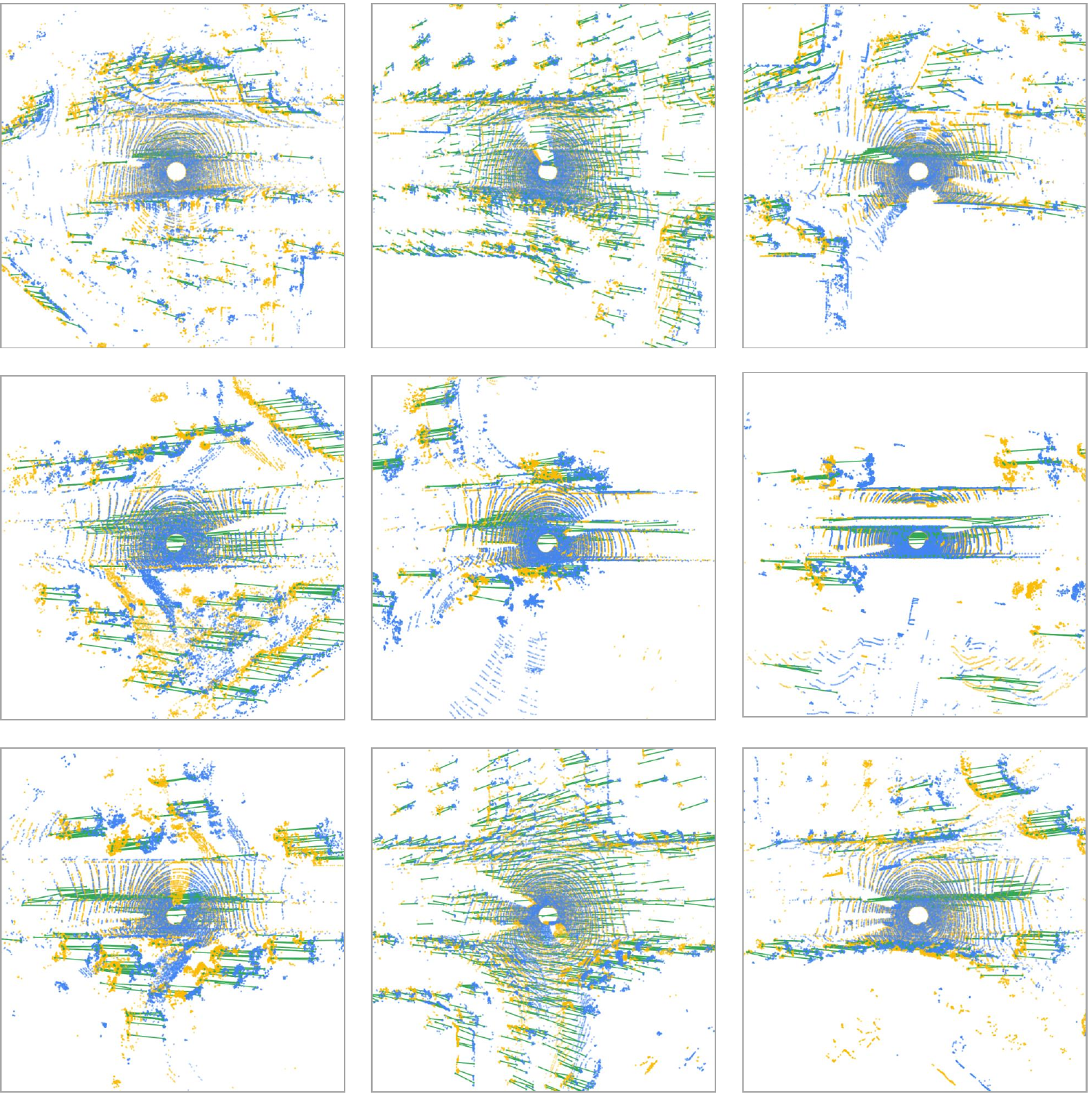}
\vspace{-4mm}
\caption{\textbf{The visualization of point cloud registration on Apollo-Southbay.} Yellow and blue points indicate input point pair frames, with green lines showing accurate correspondences (inliers) and red lines showing inaccurate ones (outliers). This figure is best viewed in zoomed-in color mode.}
\vspace{-4mm}
\label{fig:visual3}
\end{figure*}

\vspace{3pt}
\noindent\textbf{Registration Samples.} We display more visualization about the point cloud registration performance on three large-scale outdoor datasets: KITTI in Fig. \ref{fig:visual1}, nuScenes in Fig. \ref{fig:visual2}, and Apollo-Southbay in Fig. \ref{fig:visual3}. From these figures, it is clear that our proposed method accurately registers input point pairs even though there are large displacements (low-overlap problem).

\section{Limitations}
\label{sec:limit}
This paper introduces a temporally enhanced outdoor registration methods, which store memory buffers to establish long-term temporal correlations. Although this method works well in both outdoor and multiview indoor methods, its wider generalization ability on non-temporal data needs further validation. 


\end{document}